\pdfoutput=1

\documentclass[11pt]{article}

\usepackage{ACL2023}

\usepackage{times}
\usepackage{latexsym}
\usepackage{amsmath}
\usepackage{subfigure}
\usepackage{booktabs}
\usepackage{adjustbox}
\usepackage{subcaption}
\usepackage{bbding}
\usepackage{hyperref}
\usepackage[T1]{fontenc}

\usepackage[utf8]{inputenc}

\usepackage{microtype}

\usepackage{inconsolata}

\usepackage[linesnumbered,ruled,vlined,algo2e]{algorithm2e}
\usepackage{graphicx}
\usepackage{xcolor}  
\definecolor{darkgreen}{rgb}{0.0, 0.5, 0.0}

\title{Layer-Aware Task Arithmetic: Disentangling Task-Specific and Instruction-Following Knowledge}

\author{Yan-Lun Chen$^*$, Yi-Ru Wei$^*$, Chia-Yi Hsu$^*$, Chia-Mu Yu$^*$, Chun-Ying Huang$^*$, Ying-Dar Lin$^*$\\ {\bf Yu-Sung Wu$^*$ \and Wei-Bin Lee$^\spadesuit $}  \\
        $^*$ National Yang Ming Chiao Tung University\\ $^\spadesuit $ Hon Hai Research Institute  }

\begin{document}
\maketitle
\begin{abstract}
Large language models (LLMs) demonstrate strong task-specific capabilities through fine-tuning, but merging multiple fine-tuned models often leads to degraded performance due to overlapping instruction-following components. Task Arithmetic (TA), which combines task vectors derived from fine-tuning, enables multi-task learning and task forgetting but struggles to isolate task-specific knowledge from general instruction-following behavior. To address this, we propose \textit{Layer-Aware Task Arithmetic (LATA)}, a novel approach that assigns layer-specific weights to task vectors based on their alignment with instruction-following or task-specific components. By amplifying task-relevant layers and attenuating instruction-following layers, LATA improves task learning and forgetting performance while preserving overall model utility. Experiments on multiple benchmarks, including WikiText-2, GSM8K, and HumanEval, demonstrate that LATA outperforms existing methods in both multi-task learning and selective task forgetting, achieving higher task accuracy and alignment with minimal degradation in output quality. Our findings highlight the importance of layer-wise analysis in disentangling task-specific and general-purpose knowledge, offering a robust framework for efficient model merging and editing.
\end{abstract}

\section{Introduction}\label{sec: Introduction}
Existing large language models (LLMs) demonstrate robust conversational abilities but often require fine-tuning on specialized datasets for optimal task performance. Model merging combines multiple fine-tuned models into a single multi-task system. A common approach is \textit{task arithmetic} (TA) \cite{ilharco2023}, which adds or subtracts parameter differences (\textit{task vectors}) obtained before and after fine-tuning. By manipulating these vectors, TA enables a model to gain or discard specific task capabilities.

\begin{figure}[t]
\centering
\subfigure[Task vectors encode both instruction-following and task-specific capabilities.]{
\label{fig:1-1}
\includegraphics[width=0.2\textwidth]{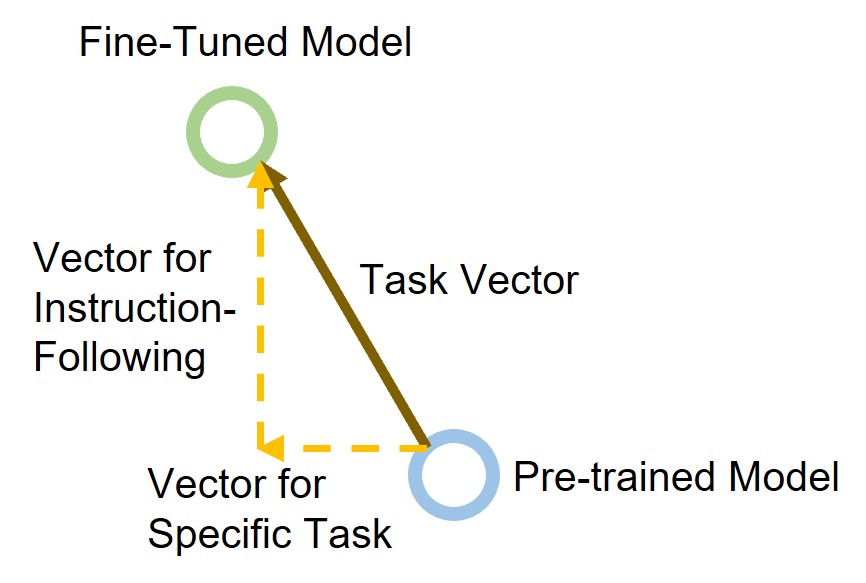}}\quad\quad\subfigure[Overlapping instruction-following components degrade merged model performance.]{
\label{fig:1-2}
\includegraphics[width=0.21\textwidth]{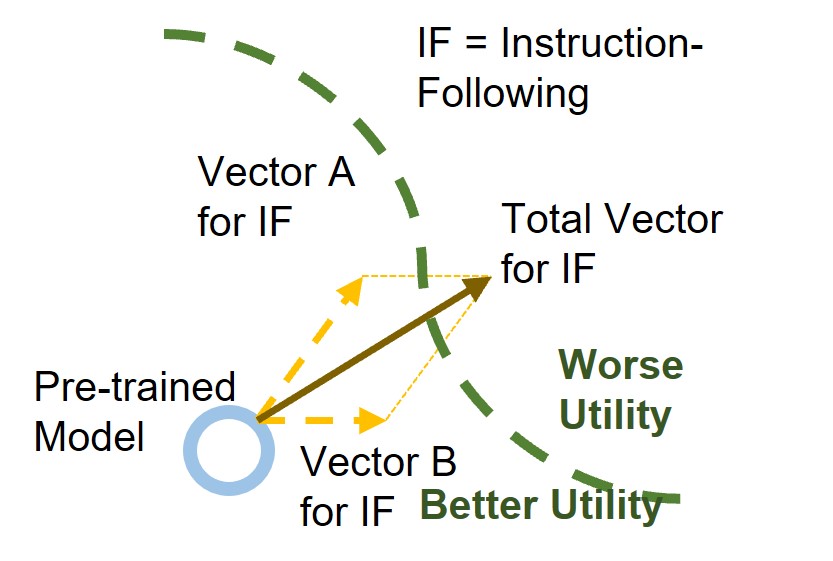}}
\caption{Challenges in task arithmetic, highlighting interference between instruction-following and task-specific components.}
\label{fig:1}
\end{figure}

Models fine-tuned for specific tasks typically stem from instruction-following LLMs \cite{dodge2020fine}. During fine-tuning, instruction-following behavior is further reinforced alongside the target task capability (Figure~\ref{fig:1-1}). Consequently, each task vector encodes both instruction-following and task-specific components. Merging multiple task vectors via TA can introduce overlapping instruction-following components, leading to worse utility in the merged model (Figure~\ref{fig:1-2}) and degrading overall output quality. Moreover, overlapping parameters across different tasks may lower performance on individual tasks when tasks are merged together.

To mitigate negative effects from overlapping instruction-following components, one must discard those portions of the task vectors and preserve only the segments that emphasize the target task. However, effectively isolating task-oriented segments remains an open challenge.

In TA, the direction of a task vector determines how the target model’s capabilities shift. We can view the full vector as a collection of layer-specific vectors, one per layer. Comparing each layer's vector with that of an instruction-following model reveals whether it focuses on instruction-following (high similarity) or on the specific task (low similarity).

Based on this observation, we propose \emph{Layer-Aware Task Arithmetic} (LATA), which assigns different weights to each layer of the task vector. Layers aligned with the target task receive larger weights, amplifying their effect on the final model, while layers emphasizing instruction-following receive smaller weights (or are disregarded) to reduce negative impact.

Our experiments show that LATA not only preserves output quality in task learning but also achieves better overall performance on each task than existing approaches. In task forgetting (the subtractive operation in TA), LATA likewise demonstrates strong effectiveness, selectively removing capabilities with minimal overall degradation.

\paragraph{Contribution} We introduce LATA, an approach to selectively amplify task-specific segments within a task vector and suppress overlapping instruction-following components. LATA preserves the merged model’s quality and achieves higher performance on multiple tasks compared to previous methods. LATA also excels at selectively removing undesired capabilities, incurring minimal harm to the model’s remaining skills.

\section{Related Work}\label{sec: Related Work}
Combining model capabilities without additional training has attracted growing attention. Model merging fuses weights of separately fine-tuned models for multi-task learning~\cite{choi2024}, and simple averaging can improve accuracy and robustness~\cite{wortsman2022}. TIES~\cite{yadav2023} resets negligible changes to address sign conflicts, reducing performance drops; Delta-sparsification (DARE)~\cite{yu2024} discards up to 99\% of fine-tuning deltas to merge multiple homologous models. Most research aims to minimize utility loss of merged LLMs~\cite{matena2022,jin2023dataless,zhou-etal-2024-metagpt,guodong24neurips,lu2024,dai2025leveraging,lai2025mediatormemoryefficientllmmerging}, while \citet{yang2024adamerging,yang2024representation,bowen2024taskvectorsselectivetask,gargiulo2025tasksingularvectorsreducing} explore merging computer vision models using key parts of task vectors.

An alternative line of research, \emph{task arithmetic}~(\textit{TA}), views tasks as weight update vectors composed via vector operations. \citet{ilharco2023} define a \emph{task vector} as the difference between a fine-tuned model and its base, enabling multiple tasks to be learned simultaneously and new tasks to be inferred without retraining. Negating a task vector selectively unlearns a specific task with minimal impact on others, implying that model weights shift independently per task. TA has been considered in fine-tuning~\citep{zhang2023composing,choi2024} and alignment~\citep{zhao-etal-2024-defending-large,li2025safety,hazra-etal-2024-safety} contexts.

In this paper, we focus on TA for both task learning and forgetting. Existing methods generally merge or edit entire models without distinguishing which layers encode task-specific versus general knowledge. In contrast, our proposed LATA performs a \emph{layer-wise analysis} to separate generic utility from task-specific effects, enabling selective amplification or removal of tasks while preserving overall performance.


\section{Background Knowledge}\label{sec: Background Knowledge}
Given $\theta_{\text{pre}}$ as the weights of a pre-trained LLM and $\theta_{\text{ft}}$ as the parameters of the LLM fine-tuned for a target task, TA \cite{ilharco2023} proposes the following formula to obtain the task vector $\tau$:  
\begin{align}  
\tau = \theta_{\text{ft}} - \theta_{\text{pre}}  
\end{align}  
where $\tau$ represents the task vector for the target task, indicating the model's capability to perform the target task.  

TA further proposes that task vectors for different target tasks can be added to a single model, enabling the model to simultaneously perform multiple target tasks. This achieves the effect of \textit{task learning}:  
\begin{align}  
\theta_{\text{merged}} = \theta_{\text{target}} + \sum_{i=1}^{t} \lambda_i \tau_i  
\end{align}  
where $t$ is the total number of target tasks, $\lambda_i$ is a scaling coefficient for the vector, $\theta_{\text{target}}$ is the original parameters of the target model, and $\theta_{\text{merged}}$ is the model after merging via TA. The merged model can simultaneously improve its performance on multiple target tasks.  

On the other hand, in the \textit{task forgetting}, the task vector can also be used to remove the model's ability for specific tasks:  
\begin{align}  
\theta_{\text{unable}} = \theta_{\text{able}} - \lambda \tau  
\end{align}  
Here, $\tau$ represents the task vector for the task to be removed. After subtracting the task vector, the model's performance on the removed task will decrease.

\begin{figure}[t]
    \centering    \includegraphics[width=0.47\textwidth]{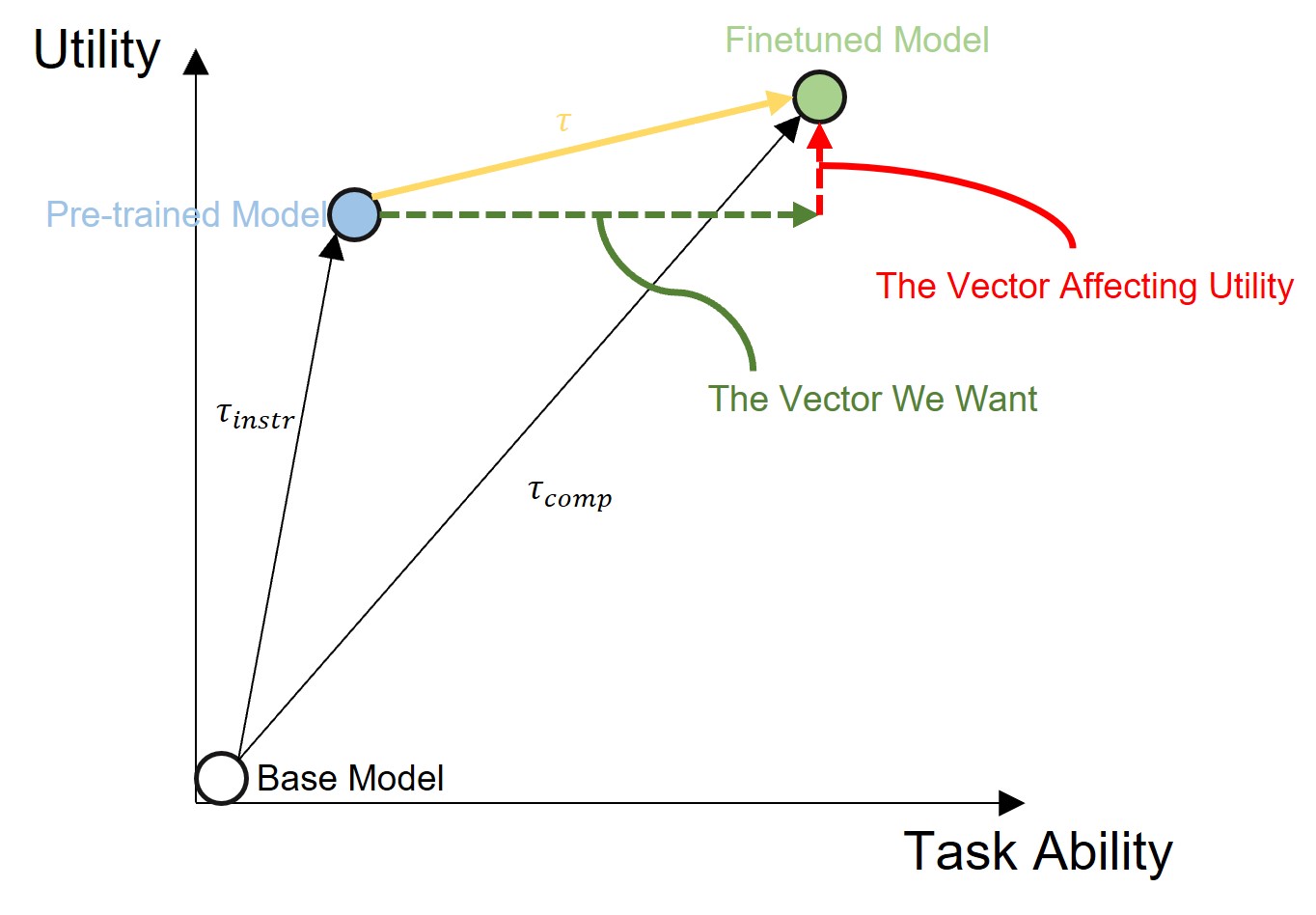}
    \caption{The difference between instruction, complex and task vector. In LATA, we emphasize extracting and applying more \textcolor{darkgreen}{green} vectors that positively impact the target task, while minimizing \textcolor{red}{red} vectors that could degrade the merged model’s utility.}
    \label{fig:2}
\end{figure}

\begin{figure*}[t]
    \centering    \includegraphics[width=0.9\textwidth]{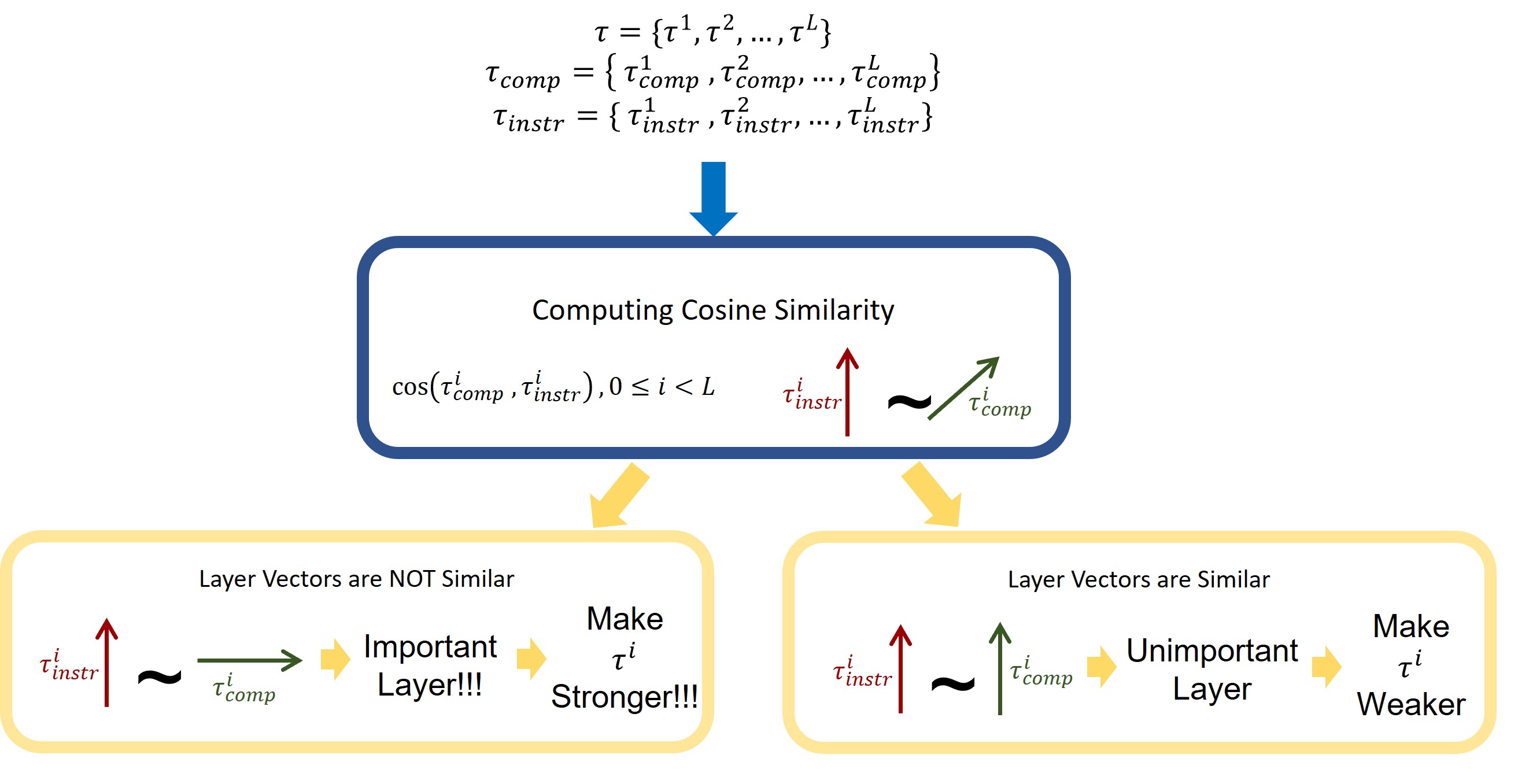}
    \caption{\textbf{Method for identifying important layers:} We compute the cosine similarity of each layer vector between the complex vector and the instruction vector. Layers with lower similarity are less related to instruction-following and likely enhance the target task, so we strengthen them. Conversely, layers with higher similarity align more with instruction-following and have lower task relevance, so we attenuate them to reduce their impact on utility.}
    \label{fig:3}
\end{figure*}

\section{Proposed Method}
Here, we present our proposed method, \textit{Layer-Aware Task Arithmetic} (LATA). First, we define a \textit{base model} as a model that does not possess instruction-following capabilities, such as Llama-3-8B \cite{grattafiori2024llama3herdmodels}. We also define a \textit{pre-trained model} as a model with instruction-following capabilities, such as Llama-3-8B-Instruct \cite{grattafiori2024llama3herdmodels}. Moreover, for multiple target tasks, we obtain models that are fine-tuned from the pre-trained model for each specific task, resulting in models tailored to their respective tasks. We refer to these models as \textit{fine-tuned models}, which are derived from the pre-trained model through fine-tuning. LATA consists of the following four steps. 

\paragraph{Step 1: Deriving Instruction Vector and Complex Vector}
We define the \textit{instruction vector} by subtracting the base model’s parameters from the pre-trained model’s parameters:

\begin{align}
\tau_{\text{instr}} = \theta_{\text{pre}} - \theta_{\text{base}}.
\end{align}

This captures the instruction-following capability. We then define the \textit{complex vector} by subtracting the base model’s parameters from those of each fine-tuned model:

\begin{align}
\tau_{\text{comp}} = \theta_{\text{ft}} - \theta_{\text{base}}.
\end{align}

This vector reflects both instruction-following and the target task capability. Figure~\ref{fig:2} shows how we obtain the instruction and complex vectors.

\paragraph{Step 2: Computing Layerwise Similarity}
We split the instruction and complex vectors into \textit{layer vectors}, with each layer’s parameters forming a small vector. Thus, the complete task vector is $\tau = \{\tau^1, \tau^2, \dots, \tau^L\}$, where $L$ is the number of layers.

To isolate target-task elements in the complex vector from instruction-following elements, we compute the cosine similarity between the instruction and complex vectors at each layer:
\begin{align}
\cos(\tau^i_{\text{comp}}, \tau^i_{\text{instr}}), \quad 0 \leq i < L.
\end{align}
Figure~\ref{fig:3} illustrates that layers showing higher similarity primarily capture instruction-following capabilities. Assigning smaller weights to these layers during TA reduces their impact on the merged model, preserving instruction-following quality. In contrast, layers with lower similarity have less effect on instruction following, so we assign them greater weights to boost target-task performance while maintaining overall utility.

\paragraph{Step 3: Deriving Pure Vector}
We obtain the target-task vector $\tau$ by subtracting the pre-trained model’s parameters from the fine-tuned model’s parameters:

\begin{align}
\tau = \theta_{\text{ft}} - \theta_{\text{pre}}.
\end{align}

Next, we split $\tau$ into layer vectors and compute each layer’s cosine similarity to the instruction and complex vectors. Layers with higher similarity receive smaller weights, and those with lower similarity receive larger weights. The resulting weighted vector is called the \emph{pure vector} because it preserves the task’s core functionality. We propose three approaches to obtain this pure vector $\tau'$.

\begin{enumerate}
    \item \textbf{Linear-Drop-by-Rank:} We rank each layer by its cosine similarity between the complex and instruction vectors, then assign weights from 0 to 1 based on rank:
   \begin{align}
   \tau' = \{\tau^{i'} \mid \tau^{i'} = \tfrac{r_i}{L}\,\tau^i,\; 1 \le r_i \le L\}
   \end{align}
   Here, $r_i$ is the rank, and higher ranks receive larger weights, indicating greater emphasis on the target task.

    \item \textbf{Logarithmic-Drop-by-Rank:} Similar to Linear-Drop-by-Rank, but because of the correlation between layers, we use a logarithmic curve:
    
    \begin{small}
   \begin{align}
   \tau' = \{\tau^{i'} \mid \tau^{i'} = \log_L(r_i)\,\tau^i,\; 1 \le r_i \le L\}.
   \end{align}
   \end{small}

   This reduces weight differences among higher-ranked layers, better reflecting inter-layer correlations in some architectures.

    \item \textbf{Drop-with-Threshold:} We set a threshold $\sigma$. If the cosine similarity of a layer exceeds $\sigma$, that layer’s vector is dropped (set to zero); otherwise, it is kept:
   \resizebox{\linewidth}{!}{
\centering
\begin{minipage}{1.25\linewidth}
\begin{align}  
\tau' = \begin{Bmatrix} \tau^{i'} {\Big|} \tau^{i'} =   
\begin{cases}   
\tau^i, & \cos(\tau^i_{\text{comp}}, \tau^i_{\text{instr}}) < \sigma \\
0, & \cos(\tau^i_{\text{comp}}, \tau^i_{\text{instr}}) \geq \sigma  
\end{cases} \end{Bmatrix}
\end{align}  
\end{minipage}
}
   This approach is useful when only a small subset of layers significantly affects the target task. By focusing on these layers, we enhance task performance.
\end{enumerate}

\paragraph{Step 4: Performing TA with Pure Vector}
Through LATA, we can obtain multiple distinct pure vectors for different target tasks. These vectors are then added to a target model via TA:

\begin{align}
\theta'_{\text{merged}} 
= \theta_{\text{target}} 
+ \sum_{i=1}^t \lambda_i \tau_i',
\end{align}
where $\lambda_i$ is the scaling coefficient for each pure vector $\tau_i'$. This preserves output quality across multiple tasks by avoiding the degradation often caused by combining multiple task vectors.

Similarly, these pure vectors can be used to remove specific capabilities:
\begin{align}
\theta'_{\text{unable}} 
= \theta_{\text{able}} 
- \lambda' \tau',
\end{align}where $\lambda'$ is the scaling coefficient for the pure vector $\tau'$. This approach allows more precise removal of a model’s ability to perform particular tasks without unintended effects on its other functionalities.

\section{Evaluation}
We conduct two experiments. The first is the \textit{task learning} scenario (see Section~\ref{sec: Related Work}), merging three target tasks (unalignment, math, and code) into a single model via TA’s additive operation. The second is the \textit{task forgetting} scenario (see Section~\ref{sec: Related Work}), using TA’s subtractive operation to reduce harmful content and improve alignment \cite{ilharco2023, bhardwaj-etal-2024-language}.

All experiments were conducted on an NVIDIA H200 GPU with 141GB of memory and dual Intel® Xeon® Platinum 8480C processors (112 cores, 2.00–3.80 GHz).

\subsection{Setup}
\paragraph{Dataset}
For task learning, we use WikiText-2 \cite{merity2017pointer} to evaluate the utility of the merged model’s outputs. For the unalignment (UA) task, we adopt the dataset designed by \citet{qi2024finetuning}, which includes 11 harmful categories defined in the usage policies of OpenAI and Llama 2, each category containing 30 harmful questions. 
We use GSM8K \cite{Cobbe2021TrainingVT} to assess the model’s math capability. 
For code generation, we employ HumanEval \cite{chen2021evaluating} as our evaluation metric.

For task forgetting, we also used the same question dataset \cite{qi2024finetuning} of 11 harmful categories from the UA (unalignment) task for model testing. 
Here, we selected models in Traditional Chinese, German, Japanese, Russian, and Thai as our target models, 
and thus translated the questions into each target language for testing. 
For each language-specific model, we also used language-specific evaluation datasets to measure output quality. 
We employed TMMLU+ \cite{tam2024tmmlu} to evaluate the Traditional Chinese model; 
JAQKET\_v2~\cite{suzuki2020jaqket}, JSQuAD, and JCommonsenseQA~\cite{kurihara-etal-2022-jglue} for the Japanese model; 
German / Russian SQuAD \cite{Artetxe_2020}, TruthfulQA \cite{lai-etal-2023-okapi}, and NLI \cite{conneau-etal-2018-xnli} for the German and Russian models, respectively; 
and Thai SQuAD \cite{Artetxe_2020} and NLI \cite{conneau-etal-2018-xnli} for the Thai model.

\paragraph{Model} We used Gemma-2-9b~\cite{gemmateam2024gemma2improvingopen} and Llama-3-8B~\cite{grattafiori2024llama3herdmodels} to evaluate LATA, with both models serving as base models and Gemma-2-9B-it and Llama-3-8B-Instruct as pre-trained or target models. Table \ref{table:1} presents the fine-tuned models for task learning. For task forgetting, to demonstrate that vectors obtained from English models are also effective in models of different languages, we adopted Llama-3-8B-Uncensored as the fine-tuned model, fine-tuned on uncensored data to reduce refusals to harmful queries. In addition, five language-specific versions, trained on their respective target languages but not heavily aligned, were used as target models listed in Table \ref{table:2}. More details of each model used for task learning/forgetting are provided in Appendix~\ref{appdix:models}.

\begin{table}[]
\centering
\begin{adjustbox}{max width=.49\textwidth}
\begin{tabular}{@{}c|c|c@{}}
\hline
\centering
Architecture & Gemma-2-9b & Llama-3-8b \\
\hline
UA & gemma-2-9b-it-abliterated & DevsDoCode/LLama-3-8b-Uncensored \\
\hline
Math & kyungeun/gemma-2-9b-it-mathinstruct & TIGER-Lab/MAmmoTH2-8B-Plus \\
\hline
Code & TeamDelta/gemma\_coder\_9b & budecosystem/code-millenials-8b\\
\hline
\end{tabular}
\end{adjustbox}
\caption{Fine-tuned models for task learning.}\label{table:1}
\end{table}

\begin{table}[t]\small
\centering
\begin{tabular}{@{}c|c@{}}
\hline
Language & Target model \\ 
\hline
Chinese (zh-tw) & Llama3-TAIDE-LX-8B-Chat-Alpha1 \\
\hline
Japanese & Llama3-DiscoLeo-Instruct-8B-v0.1 \\
\hline
German & Llama-3-ELYZA-JP-8B \\
\hline
Russian & saiga\_llama3\_8b \\
\hline
Thai & llama-3-typhoon-v1.5-8b-instruct \\ 
\hline
\end{tabular}
\caption{Target models for task forgetting.}
\label{table:2}
\end{table}

\paragraph{Metric} We use the following metrics for evaluation.
\begin{enumerate}
    \item \textbf{Utility} We use WikiText-2 Benchmark\footnote{\href{https://github.com/EleutherAI/lm-evaluation-harness}{https://github.com/EleutherAI/lm-evaluation-harness}\label{lm-eval}} \cite{merity2017pointer} to compute the perplexity of the merged model to examine the issue of quality degradation in the model's output. For models in different languages, we use different metrics to evaluate their capabilities:
\begin{enumerate}
    \item \textbf{Traditional Chinese} We use TMMLU+\textsuperscript{\ref{lm-eval}} for evaluation. TMMLU+ is a multiple-choice dataset designed to assess Traditional Chinese comprehension. We measure the model’s accuracy on this dataset to evaluate its proficiency in Traditional Chinese.
    \item \textbf{Japanese} We evaluate the model using exact-match score for JAQKET\_v2\textsuperscript{\ref{lm-eval}} and JSQuAD\textsuperscript{\ref{lm-eval}}, and accuracy for JCommonsenseQA\textsuperscript{\ref{lm-eval}}. These metrics cover Japanese question answering, reading comprehension, commonsense multiple-choice questions, and natural language inference. 
    \item \textbf{German, Russian, and Thai} We separately use the German, Russian, and Thai versions of SQuAD\textsuperscript{\ref{lm-eval}} F1-score and NLI \textsuperscript{\ref{lm-eval}} accuracy for evaluation. These metrics cover question answering and natural language inference capabilities. For the German and Russian models, we also employ the respective language versions of TruthfulQA\textsuperscript{\ref{lm-eval}} accuracy to assess their question-answering performance.
\end{enumerate}

    \item \textbf{Unalignment (UA)}
We use GPT-4 \cite{openai2024gpt4technicalreport} to score the risk level of the model's output (on a scale of 1 to 5, where higher scores indicate more unsafe outputs) \cite{qi2024finetuning}.

    \item \textbf{Math}
We evaluate the model's performance on the GSM8K\textsuperscript{\ref{lm-eval}} \cite{Cobbe2021TrainingVT} dataset using zero-shot accuracy.

    \item \textbf{Code}
We assess the model's ability to generate code using pass@1 on the HumanEval benchmark \cite{chen2021evaluating}.
\end{enumerate}

\paragraph{Baseline}
We consider the ordinary TA~\cite{ilharco2023}, TIES~\cite{yadav2023}, and DARE~\cite{yu2024} as baseline methods in task learning since these are all primarily based on TA, designed for LLMs, and do not require additional data. For task forgetting, we also consider TA, DARE, and Safety Arithmetic~\cite{hazra-etal-2024-safety}. TA has been described in Section~\ref{sec: Background Knowledge}. TIES reduces interference by retaining only the top $k \times 100\%$ of parameters (by magnitude) in the task vector. DARE tackles parameter interference by randomly dropping $p \times 100\%$ of the parameters in the task vector. Safety Arithmetic first uses $\lambda$ for harm direction removal, then applies $\alpha$ to add the in-context vector into the model to enhance alignment. We show the configuration of each baseline below.

\begin{itemize}
    \item \textbf{TA:} We follow the description in Section~\ref{sec: Background Knowledge} to implement TA. We set the scaling coefficient $\lambda$ as $0.5$ (and $1.0$) in task learning and $0.8$ in task forgetting. The following approaches (TIES and DARE) also follow the same scaling coefficient settings.

    \item \textbf{TIES:}  
    We retain the top $0.7 \times 100\%$ of parameters (by magnitude) in the task vector ($k=0.7$).  

    \item \textbf{DARE:}  
    We set the drop rate $p$ to 0.3 in both task learning and task forgetting. Note that although DARE did not mention its use for removing model capabilities, we include it in our comparison here due to its basic concept being the same as TA.  

    \item \textbf{DARE+TIES:}  
    $p=0.1$ and $k=0.9$. 
    
    \item \textbf{Safety Arithmetic} To maintain the generating capabilities of various language models, we set $\lambda=0.5$, $\alpha=0.12$ for Chinese, Russian, and Thai models, $\lambda=0.3$, $\alpha=0.12$ for German model, and $\lambda=0.3$, $\alpha=0.08$ for Japanese model.
\end{itemize}

\begin{table}[h]\tiny
\centering
\tabcolsep=1.4pt
\begin{tabular}{@{}cccccc@{}}
    \hline
    \begin{tabular}[c]{@{}c@{}}Merged\\ Tasks\end{tabular} & \begin{tabular}[c]{@{}c@{}} Merging\\ Method\end{tabular} & \begin{tabular}[c]{@{}c@{}}Utility\\ WikiText-2($\downarrow$)\end{tabular} & \begin{tabular}[c]{@{}c@{}}UA\\ GPT-4($\uparrow$)\end{tabular} & \begin{tabular}[c]{@{}c@{}}Math\\ GSM8K($\uparrow$)\end{tabular} & \begin{tabular}[c]{@{}c@{}}Code\\ HumanEval($\uparrow$)\end{tabular} \\ \midrule
    \hline
     & TA & 11.4631 & 3.7091 & 0.8211 & - \\
     & DARE & 11.6558 & 3.8000 & 0.8143 & - \\
     UA + Math& TIES & 12.3577 & 3.3303 & 0.8112 & - \\
     & DARE + TIES & 12.7110 & 3.3030 & 0.8249 & - \\
     & LATA (Ours) & \textbf{10.2726} & \textbf{3.8879} & \textbf{0.8408} & - \\
    \hline
     & TA & 10.3444 & - & 0.8347 & 0.6463 \\
     & DARE & 12.4347 & - & 0.8294 & 0.6341 \\
     Math + Code& TIES & 12.3455 & - & 0.8279 & 0.6524 \\
     & DARE + TIES & 10.4208 & - & 0.8287 & 0.6341 \\
     & LATA (Ours) & \textbf{10.2831} & - & \textbf{0.8461} & \textbf{0.6585} \\
    \hline
     & TA & 12.3533 & 3.7485 & - & 0.4878 \\
     & DARE & 12.6539 & 3.7758 & - & \textbf{0.5183} \\
     UA + Code& TIES & 12.5680 & 3.7879 & - & 0.4878 \\
     & DARE + TIES & 12.9077 & 3.5848 & - & 0.5000 \\
     & LATA (Ours) & \textbf{10.9101} & \textbf{3.8455} & - & 0.4756 \\
    \hline
     & TA & 11.8785 & 3.7576 & 0.8241 & 0.6159 \\
     & DARE & 12.3247 & 3.7152 & 0.8052 & \textbf{0.6341} \\
     UA + Math + Code& TIES & 15.7654 & 2.8727 & 0.7870 & 0.5976 \\
     & DARE + TIES & 16.9879 & 2.8061 & 0.7627 & 0.5793 \\
     & LATA (Ours) & \textbf{10.4298} & \textbf{3.7939} & \textbf{0.8431} & 0.6280 \\ 
    \hline
\end{tabular}
\caption{The performance of LATA compared with TA, DARE, TIES, and DARE+TIES (TIES applied after DARE) under Gemma-2-9b is shown for various combinations of UA, Math, and Code. We use $\lambda=1.5$ for UA and $\lambda=0.5$ for Math and Code.}\label{table:3}
\end{table}

\begin{table}[h]\tiny
\centering
\tabcolsep=0.5pt
\begin{tabular}{@{}cccccc@{}}
    \hline
    \begin{tabular}[c]{@{}c@{}}Merged\\ Tasks\end{tabular} & \begin{tabular}[c]{@{}c@{}} Merging\\ Method\end{tabular} & \begin{tabular}[c]{@{}c@{}}Utility\\ WikiText-2($\downarrow$)\end{tabular} & \begin{tabular}[c]{@{}c@{}}UA\\ GPT-4($\uparrow$)\end{tabular} & \begin{tabular}[c]{@{}c@{}}Math\\ GSM8K($\uparrow$)\end{tabular} & \begin{tabular}[c]{@{}c@{}}Code\\ HumanEval($\uparrow$)\end{tabular} \\ \midrule
    \hline
     & LATA + TIES & \textbf{10.2724} & 3.8848 & \textbf{0.8431} & - \\
     UA + Math& LATA + DARE & 10.2936 & 3.8333 & 0.8340 & - \\
     & LATA + DARE + TIES & 10.2784 & \textbf{3.9152} & 0.8324 & - \\
    \hline
     & LATA + TIES & \textbf{10.3029} & - & \textbf{0.8491} & 0.6402 \\
     Math + Code& LATA + DARE & 10.3150 & - & 0.8438 & 0.6463 \\
     & LATA + DARE + TIES & 10.3046 & - & 0.8408 & \textbf{0.6524} \\
    \hline
     & LATA + TIES & 10.9103 & 3.7970 & - & \textbf{0.4573} \\
     UA + Code& LATA + DARE & \textbf{10.9097} & \textbf{3.8394} & - & 0.4024 \\
     & LATA + DARE + TIES & 12.9204 & 3.7788 & - & 0.4512 \\
    \hline
     & LATA + TIES & \textbf{10.5050} & 3.7061 & \textbf{0.8431} & 0.6341 \\
     UA + Math + Code& LATA + DARE & 10.5219 & \textbf{3.7879} & 0.8408 & 0.6341 \\
     & LATA + DARE + TIES & 10.5137 & 3.7061 & 0.8393 & 0.6341 \\
    \hline
\end{tabular}
\caption{Results of Combining LATA with TIES, DARE, and DARE + TIES. We use $\lambda=1.5$ for UA, $\lambda=0.5$ for Math and Code. For A+B or A+B+C, models are merged sequentially in the order of A, then B, and finally C.}\label{table:4}
\end{table}

\begin{table}[t]\tiny
\centering
\tabcolsep=1.4pt
\begin{tabular}{@{}cccccc@{}}
    \hline
    \begin{tabular}[c]{@{}c@{}}Merged\\ Tasks\end{tabular} & \begin{tabular}[c]{@{}c@{}}Merging\\ Method\end{tabular} & \begin{tabular}[c]{@{}c@{}}Utility\\ WikiText-2($\downarrow$)\end{tabular} & \begin{tabular}[c]{@{}c@{}}UA\\ GPT-4($\uparrow$)\end{tabular} & \begin{tabular}[c]{@{}c@{}}Math\\ GSM8K($\uparrow$)\end{tabular} & \begin{tabular}[c]{@{}c@{}}Code\\ HumanEval($\uparrow$)\end{tabular} \\ \midrule
    \hline
     & TA & \textbf{10.0031} & \textbf{1.6212} & 0.8355 & - \\
     & DARE & 10.0146 & 1.5909 & 0.8324 & - \\
     UA + Math& TIES & 10.0167 & 1.5636 & 0.8385 & - \\
     & DARE + TIES & 10.0461 & 1.5818 & 0.8302 & - \\
     & LATA (Ours) & 10.0667 & 1.3455 & \textbf{0.8552} & - \\
    \hline
     & TA & 10.8258 & 1.6394 & - & 0.4390 \\
     & DARE & 10.8740 & \textbf{1.7061} & - & 0.4390 \\
     UA + Code& TIES & 11.8583 & 1.5121 & - & 0.4329 \\
     & DARE + TIES & 10.8553 & 1.6121 & - & \textbf{0.4512} \\
     & LATA (Ours) & \textbf{10.6848} & 1.3848 & - & 0.3902 \\
    \hline
     & TA & 10.3483 & 1.7515 & 0.8431 & 0.6463 \\
     & DARE & 10.3804 & 1.6394 & 0.8294 & 0.6463 \\
     UA + Math + Code& TIES & 10.3994 & 1.7939 & 0.8317 & \textbf{0.6585} \\
     & DARE + TIES & 10.4147 & \textbf{1.8091} & 0.8309 & 0.6524 \\
     & LATA (Ours) & \textbf{10.2860} & 1.4152 & \textbf{0.8514} & \textbf{0.6585} \\ 
    \hline
\end{tabular}
\caption{Results of task learning on Gemma-2-9b. Here, we merge models with $\lambda=0.5$ for all tasks. Since the settings and results of "Math + Code" are identical to those in Table 3, we do not repeat them here.}\label{table:5}
\end{table}

\begin{table}[h]\tiny
\centering
\tabcolsep=1.4pt
\begin{tabular}{@{}cccccc@{}}
    \hline
    \begin{tabular}[c]{@{}c@{}}Merged\\ Tasks\end{tabular} & \begin{tabular}[c]{@{}c@{}}Merging\\ Method\end{tabular} & \begin{tabular}[c]{@{}c@{}}Utility\\ WikiText-2($\downarrow$)\end{tabular} & \begin{tabular}[c]{@{}c@{}}UA\\ GPT-4($\uparrow$)\end{tabular} & \begin{tabular}[c]{@{}c@{}}Math\\ GSM8K($\uparrow$)\end{tabular} & \begin{tabular}[c]{@{}c@{}}Code\\ HumanEval($\uparrow$)\end{tabular} \\ \midrule
    \hline
     & TA & 10.7850 & \textbf{3.9758} & 0.7377 & - \\
     & DARE & 10.8753 & 3.9424 & 0.7437 & - \\
     UA + Math& TIES & 10.7638 & 3.3606 & 0.7475 & - \\
     & DARE + TIES & 10.8560 & 3.6424 & 0.7445 & - \\
     & LATA (Ours) & \textbf{10.2638} & 2.2909 & \textbf{0.8158} & - \\
    \hline
     & TA & 12.1416 & - & 0.7248 & 0.5793 \\
     & DARE & 12.4347 & - & 0.7111 & 0.5305 \\
     Math + Code& TIES & 12.3455 & - & 0.7165 & 0.5366 \\
     & DARE + TIES & 12.5877 & - & 0.6914 & 0.5061 \\
     & LATA (Ours) & \textbf{11.0208} & - & \textbf{0.8317} & \textbf{0.6280} \\
    \hline
     & TA & 11.9674 & \textbf{3.8545} & - & 0.3171 \\
     & DARE & 12.1404 & 3.8394 & - & \textbf{0.3537} \\
     UA + Code& TIES & 11.9742 & 3.3545 & - & 0.2866 \\
     & DARE + TIES & 12.0392 & 3.5061 & - & 0.2866 \\
     & LATA (Ours) & \textbf{11.2949} & 2.5667 & - & 0.3293 \\
    \hline
     & TA & 12.2611 & 3.6364 & 0.7172 & 0.5671 \\
     & DARE & 12.5596 & \textbf{3.6939} & 0.7005 & 0.5061 \\
     UA + Math + Code& TIES & 12.5602 & 3.5061 & 0.7104 & 0.5366 \\
     & DARE + TIES & 12.8658 & 3.4788 & 0.6914 & 0.5122 \\
     & LATA (Ours) & \textbf{11.0486} & 2.6394 & \textbf{0.8271} & \textbf{0.6280} \\ 
    \hline
\end{tabular}
\caption{Results of task learning on Gemma-2-9b. Here, we merge models with $\lambda=1.0$ for all tasks.}\label{table:6}
\end{table}

\begin{table}[h]\tiny
\centering
\tabcolsep=1.4pt
\begin{tabular}{@{}cccccc@{}}
    \hline
    \begin{tabular}[c]{@{}c@{}}Merged\\ Tasks\end{tabular} & \begin{tabular}[c]{@{}c@{}}Merging\\ Method\end{tabular} & \begin{tabular}[c]{@{}c@{}}Utility\\ WikiText-2($\downarrow$)\end{tabular} & \begin{tabular}[c]{@{}c@{}}UA\\ GPT-4($\uparrow$)\end{tabular} & \begin{tabular}[c]{@{}c@{}}Math\\ GSM8K($\uparrow$)\end{tabular} & \begin{tabular}[c]{@{}c@{}}Code\\ HumanEval($\uparrow$)\end{tabular} \\ \midrule
    \hline
     & TA & \textbf{9.0025} & 3.6303 & \textbf{0.8089} & - \\
     & DARE & 9.0559 & 3.5606 & 0.8074 & - \\
     UA + Math& TIES & 9.1528 & 3.3788 & 0.8036 & - \\
     & DARE + TIES & 9.0055 & 3.5515 & 0.7923 & - \\
     & LATA (Ours) & 9.3160 & \textbf{3.7667} & 0.7847 & - \\
    \hline
     & TA & 10.0648 & - & 0.6664 & \textbf{0.3415} \\
     & DARE & 10.1674 & - & 0.6558 & \textbf{0.3415} \\
     Math + Code& TIES & 10.0103 & - & 0.6914 & 0.3293 \\
     & DARE + TIES & 10.2170 & - & 0.6778 & 0.2927 \\
     & LATA (Ours) & \textbf{9.9947} & - & \textbf{0.7491} & 0.2439 \\
    \hline
     & TA & 10.4806 & \textbf{3.7879} & - & 0.2317 \\
     & DARE & 10.4840 & 3.7273 & - & 0.2256 \\
     UA + Code& TIES & \textbf{10.2491} & 3.5667 & - & 0.2256 \\
     & DARE + TIES & 10.5172 & 3.6606 & - & 0.1951 \\
     & LATA (Ours) & 10.4579 & 3.5030 & - & \textbf{0.2500} \\
    \hline
     & TA & 9.9398 & 3.6333 & 0.6626 & 0.2987 \\
     & DARE & 10.0415 & \textbf{3.8000} & 0.6732 & 0.3171 \\
     UA + Math + Code& TIES & 9.9066 & 3.5030 & 0.6793 & 0.3171 \\
     & DARE + TIES & 10.1180 & 3.6727 & 0.6634 & \textbf{0.3537} \\
     & LATA (Ours) & \textbf{9.9057} & 3.7939 & \textbf{0.7316} & 0.2378 \\ 
    \hline
\end{tabular}
\caption{Results of task learning on Llama-3-8b. Here, we merge models with $\lambda=0.5$ for all tasks.}\label{table:7}
\end{table}

\subsection{Result}\label{sec:results}
\paragraph{Task Learning}
We evaluate LATA on Gemma-2-9b (Linear-Drop-by-Rank) and Llama-3-8B (Logarithmic-Drop-by-Rank) with scaling coefficients set to $0.5$ and $1.0$. Table~\ref{table:3} shows results under Gemma-2-9b. Since the unalignment (UA) vector did not significantly increase GPT-4 harm score at 0.5 or 1.0, we use a coefficient of 1.5 for UA and 0.5 for the other two tasks. Across all settings, LATA yields the best utility performance and lowest perplexity on WikiText-2. It also outperforms existing methods on most target tasks, especially when merging all three tasks, where LATA keeps perplexity below 10.5 while all others exceed 11.5. 

Compared to the Table~\ref{table:3}, where the scaling coefficient $\lambda$'s for different tasks are particularly set, Tables~\ref{table:5} and \ref{table:6} show results for coefficients 0.5 and 1.0. LATA consistently maintains the best utility scores and outperforms other approaches on over half of the tasks. Although performance in utility, math, and code slightly declines at 1.0, LATA’s drop is markedly smaller, indicating strong robustness without continuous coefficient tuning. On the other hand, to show the influences of different hyperparameters of each baseline, we perform the results in Appendix~\ref{appdix:results-gemma}.

We also investigate whether LATA can enhance DARE and TIES. Table~\ref{table:4} shows that combining LATA with these methods often yields superior utility. However, LATA+DARE+TIES typically underperforms LATA+DARE or LATA+TIES alone, mirroring the observation that DATA+TIES is weaker than DARE or TIES. Moreover, in most cases, these three-method combinations in Table~\ref{table:4} are worse than LATA alone (Table~\ref{table:3}), as TIES and DARE may zero out crucial layer vectors selected by LATA. Hence, using LATA by itself remains the best choice. 

Table~\ref{table:7} presents results under Llama-3-8B and additional results with different hyperparameters for each baseline can be found in Appendix~\ref{appdix:results-llama}. Owing to its smaller size, we adopt Logarithmic-Drop-by-Rank to account for higher interdependence among layers. LATA still sustains superior overall utility while achieving competitive or best scores in several tasks. This demonstrates LATA’s effectiveness across different architectures. In summary, LATA consistently shows clear advantages in merging multiple models.

\begin{figure}[t]
    \centering
    \subfigure[Chinese model's utility]{ \label{fig:4-1}
    \includegraphics[width=0.23\textwidth]{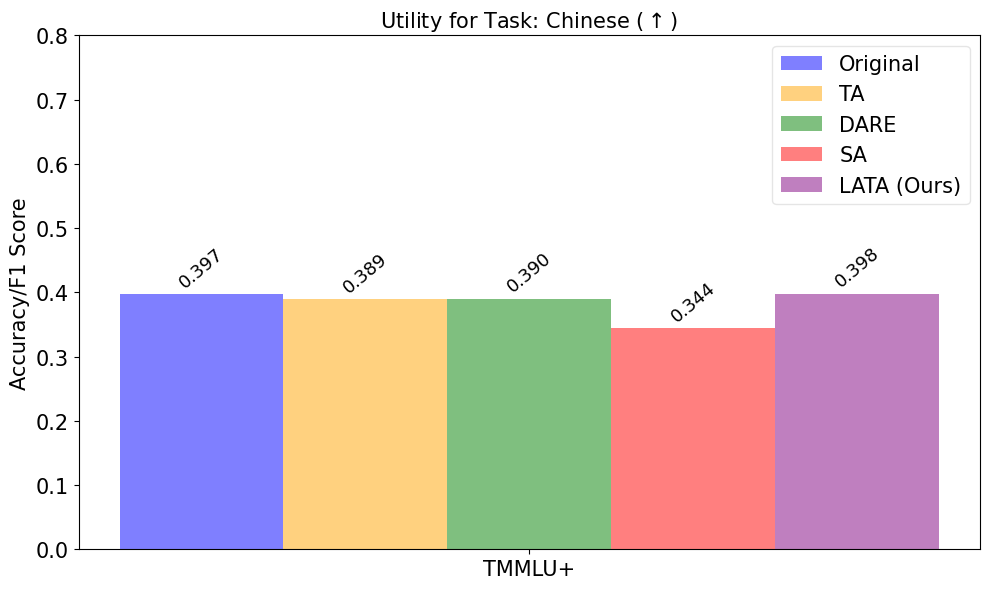}}
    \subfigure[Chinese model's toxicity]{ \label{fig:4-2}
    \includegraphics[width=0.23\textwidth]{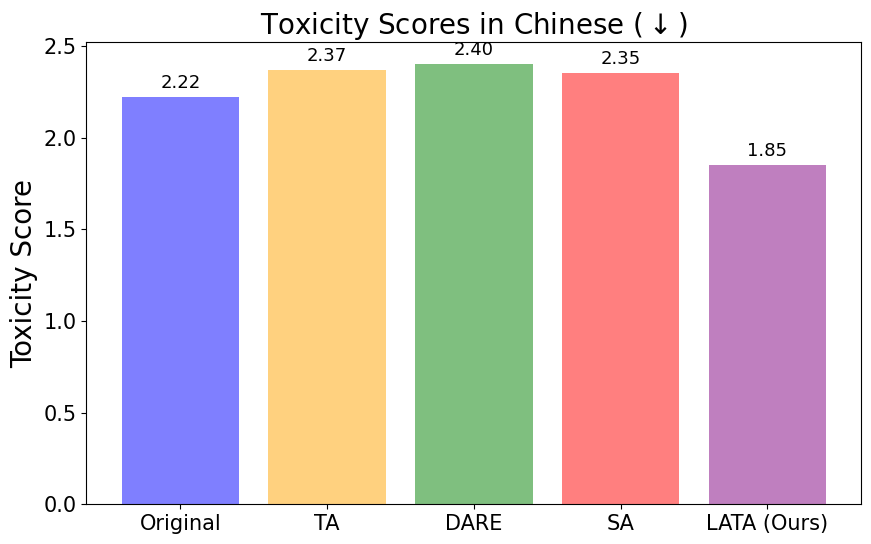}}
    \subfigure[Japanese model's utility]{ \label{fig:4-3}
    \includegraphics[width=0.23\textwidth]{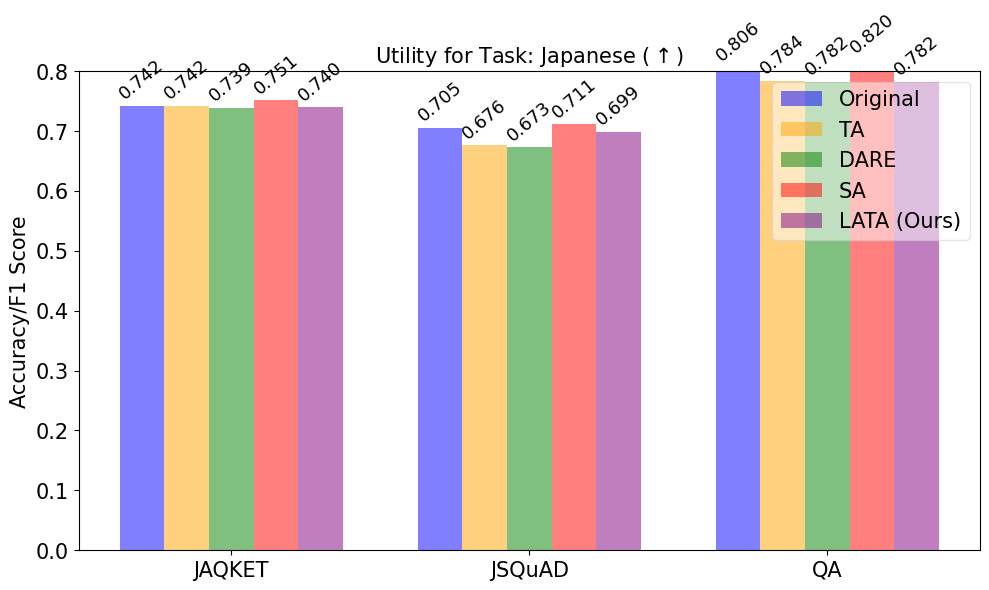}}
    \subfigure[Japanese model's toxicity]{ \label{fig:4-4}
    \includegraphics[width=0.23\textwidth]{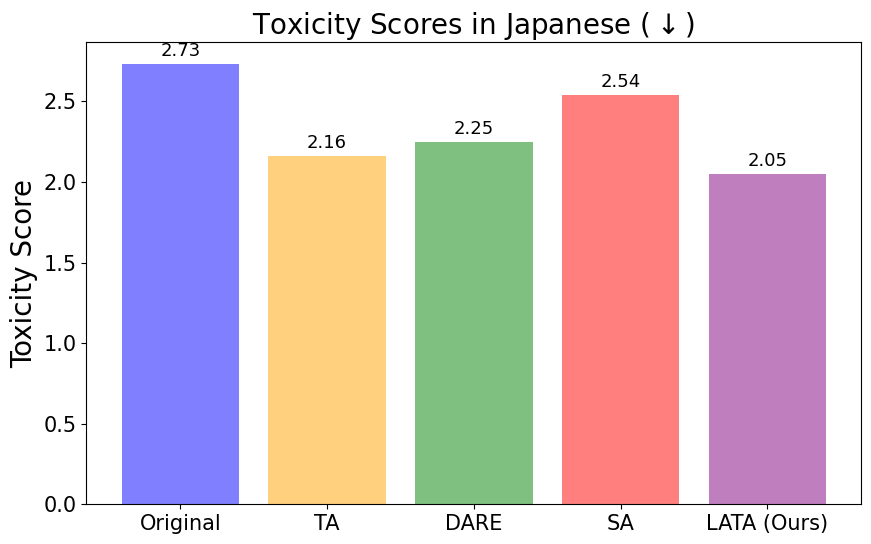}}
    \subfigure[German model's utility]{ \label{fig:4-5}
    \includegraphics[width=0.23\textwidth]{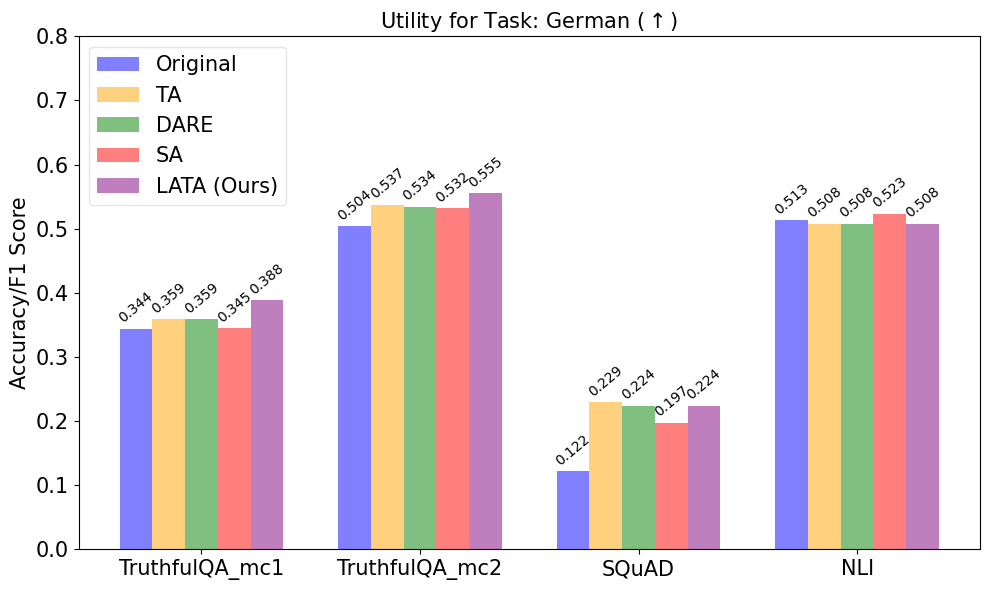}}
    \subfigure[German model's toxicity]{ \label{fig:4-6}
    \includegraphics[width=0.23\textwidth]{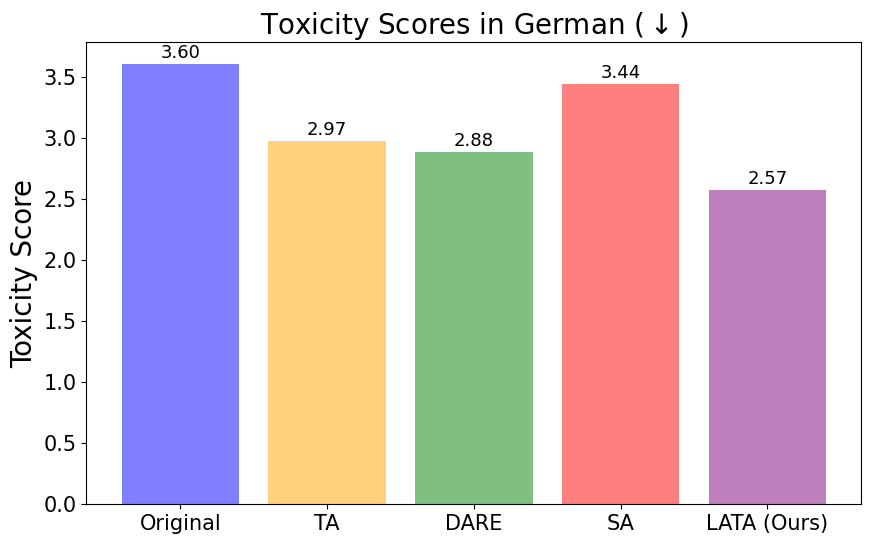}}
    \subfigure[Russian model's utility]{ \label{fig:4-7}
    \includegraphics[width=0.23\textwidth]{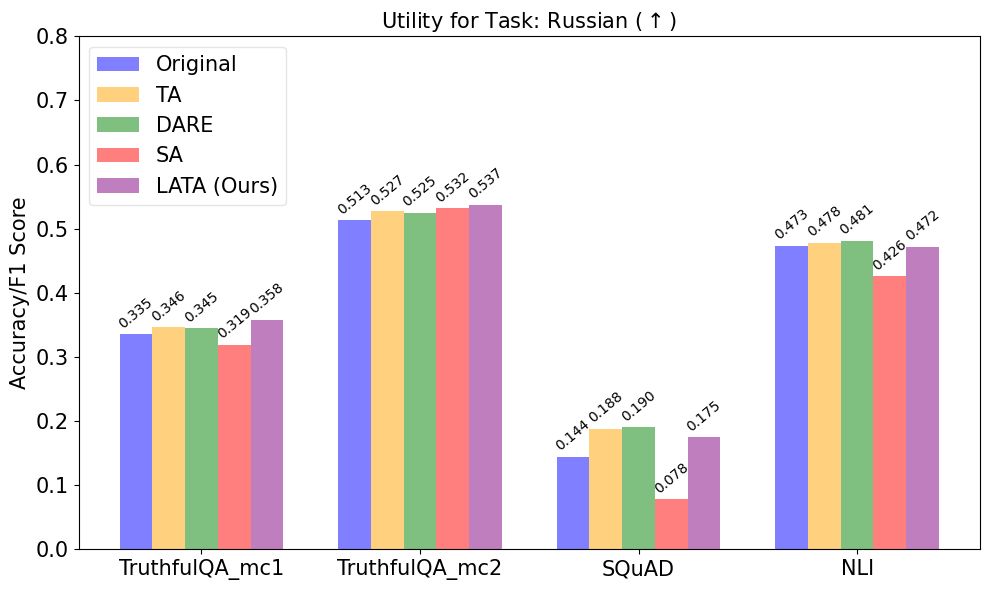}}
    \subfigure[Russian model's toxicity]{ \label{fig:4-8}
    \includegraphics[width=0.23\textwidth]{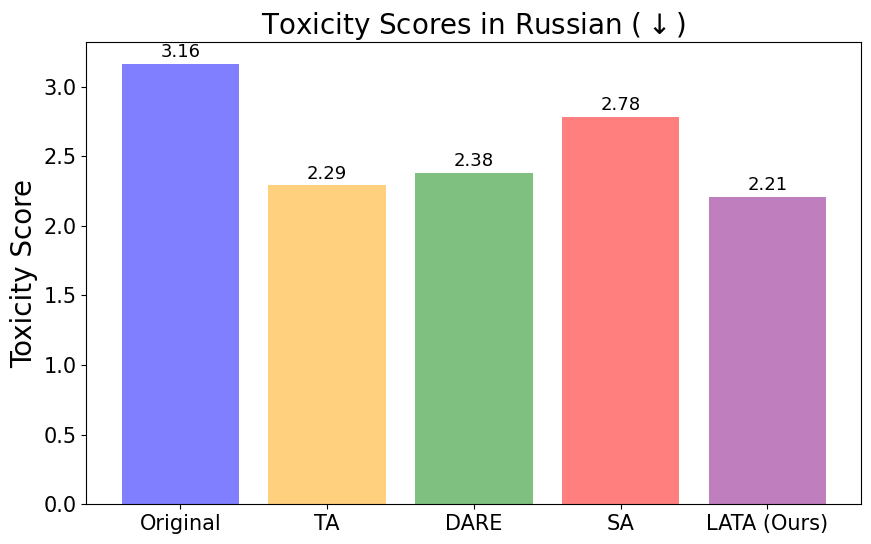}}
    \subfigure[Thai model's utility]{ \label{fig:4-9}
    \includegraphics[width=0.23\textwidth]{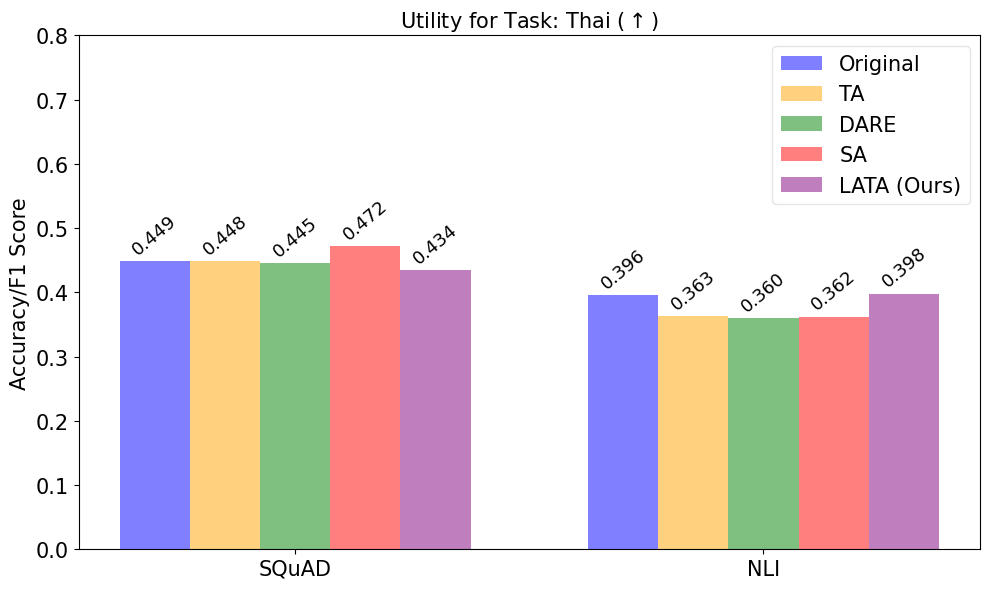}}
    \subfigure[Thai model's toxicity]{ \label{fig:4-10}
    \includegraphics[width=0.23\textwidth]{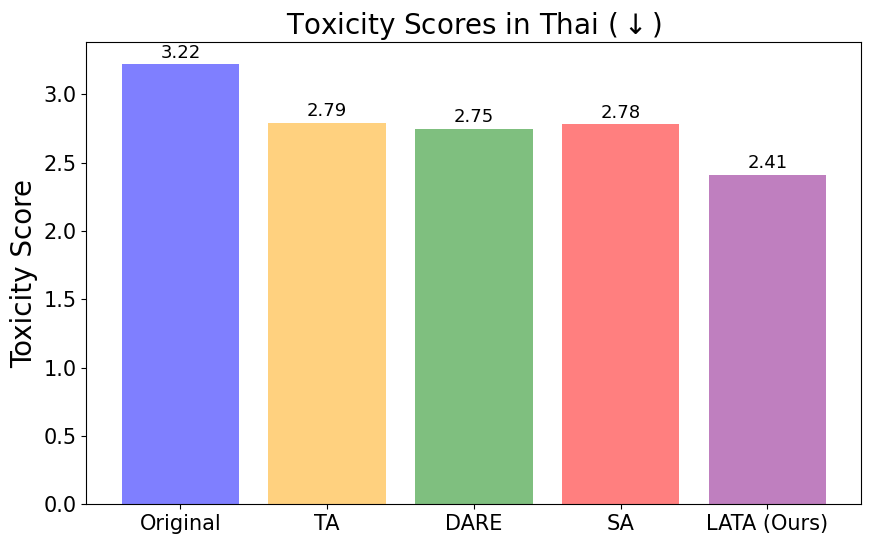}}
    \caption{Result of task forgetting}
    \label{fig:4}
\end{figure}

\paragraph{Task Forgetting}
We set the scaling coefficient $\lambda$ to $1.0$ and use Drop-with-Threshold at the threshold $\sigma$ of $0.95$ (see more discussion in Section~\ref{sec: Discussion}). Figure~\ref{fig:4} shows that applying TA’s subtractive operation to reduce harmful content substantially improves alignment. LATA consistently outperforms existing methods, reducing GPT-4 harm scores below 2 for all tested languages, notably from 3.60 to 2.57 in German. Meanwhile, utility remains on par with the original model. These results suggest LATA precisely targets task vectors for removal and, in some cases (see Section~\ref{sec: Discussion}), adjusting a minimal subset of parameters is sufficient to eliminate specific capabilities.

\section{Discussion}\label{sec: Discussion}
\begin{figure}[t]
    \centering    \includegraphics[width=0.48\textwidth]{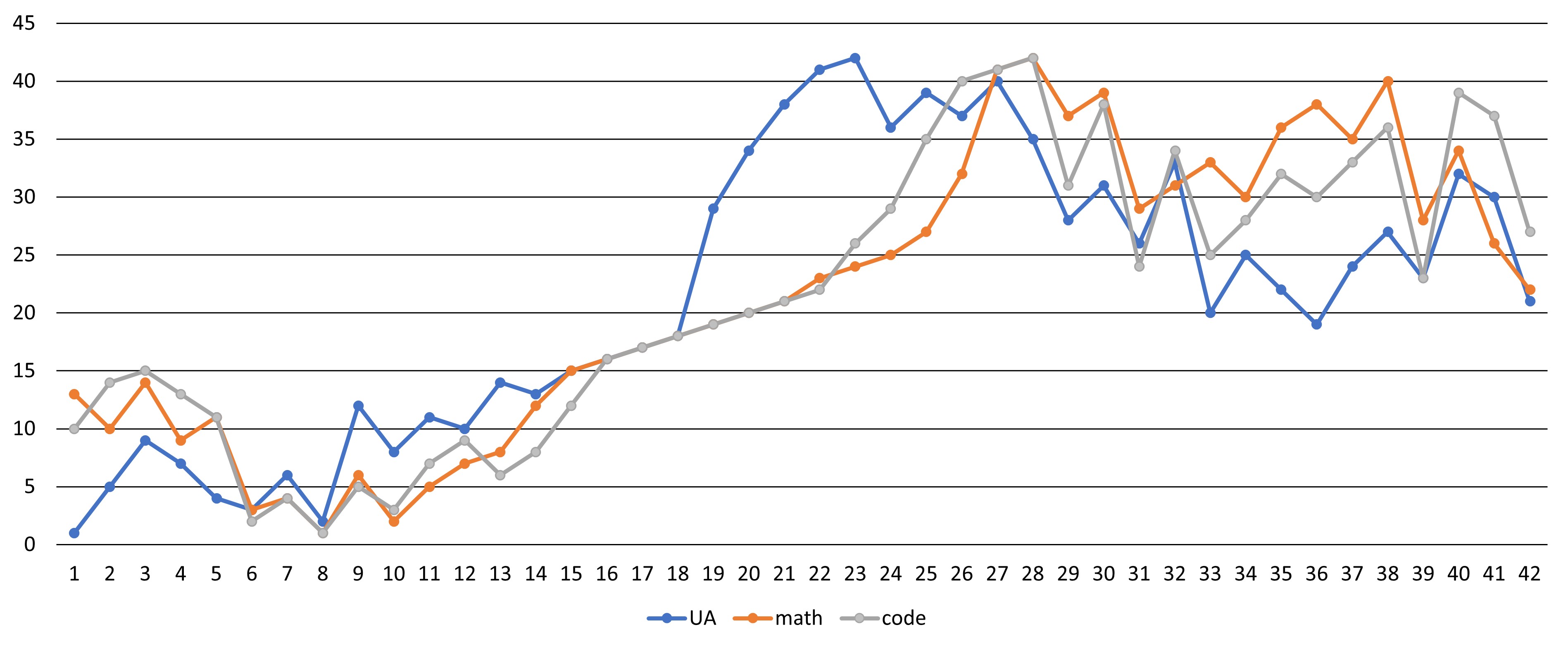}
    \caption{The graph illustrates the similarity rankings among layer vectors, with the x-axis representing the layer number and the y-axis indicating the similarity rank.}
    \label{fig:5}
\end{figure}

\paragraph{Distribution of Important Layers for Target Tasks}
Figure~\ref{fig:5} shows layer-wise similarity rankings between the three target tasks’ complex vectors for Gemma-2-9b and the instruction vector. The vertical axis is the similarity ranking, and the horizontal axis is the layer index. Layers with lower similarity (thus more impact on the target task) generally appear after layer 20, especially between layers 26 and 30. We hypothesize that earlier layers focus more on processing the input instructions, making them closer to the instruction vector and less crucial to the target task. Conversely, later layers generate outputs based on the earlier layers’ interpretations, causing parameter changes there to have greater impact on the target task and thus lower similarity with the instruction vector.

Another notable observation is the significant overlap in similarity rankings for math and code tasks. We suspect a strong intrinsic similarity between these two tasks, reflected in our experiments: when merging them simultaneously (\emph{math + code}, \emph{UA + math + code}), both tasks outperform their single-task scenarios (\emph{UA + math}, \emph{UA + code}), particularly for code. This suggests that when task vectors share substantial similarity, merging them concurrently can further enhance the resulting model’s performance on each individual task.

\paragraph{Significant Impact from a Small Fraction of Layer Vectors}
In our task forgetting experiment, we set the threshold $\sigma$ to $0.95$, meaning that layer vectors with similarities above $0.95$ were discarded. We arrived at this threshold because we observed extremely high similarity between the complex and instruction vectors for each layer, with only a handful of layer vectors showing similarity below $0.9$. Even with the threshold fixed at $0.95$, only about $10\%$ of the layer vectors were retained as \emph{pure vectors}, while the remaining $90\%$ had similarities greater than 0.95. Under the DARE concept, discarding $90\%$ of the vectors would ordinarily require rescaling the remaining $10\%$ by a factor of $\frac{1}{1 - 0.9}$ (i.e., $10\times$). However, we merely applied $\lambda=1.0$ to slightly increase these vectors, already achieving performance surpassing that of the original TA method. This finding indicates that a complete task vector indeed contains a subset of parameters that are highly critical to the target task, while a substantial portion is less significant. 
LATA successfully isolates these crucial and non-crucial segments from the task vector.

\section{Conclusion}
In this work, we introduced a novel approach (LATA) to TA, demonstrating its effectiveness in merging and fine-tuning LLMs across diverse tasks. LATA leverages dynamic task representations to achieve improved alignment and utility without compromising model performance. Through extensive experiments on benchmark datasets such as WikiText-2, GSM8K, and HumanEval, we showed that our approach consistently outperforms existing methods like DARE and TIES in balancing task-specific performance and generalization. Notably, our framework enables efficient model merging while mitigating interference between tasks, as evidenced by superior results in multi-task scenarios. Our findings highlight the potential of TA as a scalable and adaptable solution for optimizing LLMs in multi-task and cross-lingual settings. 

\clearpage
\newpage

\section*{Limitations}
LATA relies on task arithmetic, so all models must share the same architecture (identical hidden dimensions and layer structures), which limits cross-family applications. Moreover, improper scaling coefficients of task vectors ($\lambda$) can lead to instability, potentially degrading model performance or causing catastrophic forgetting.

\nocite{huang-etal-2024-chat}
\nocite{goddard-etal-2024-arcees}
\nocite{pipatanakul2023typhoon}
\nocite{elyzallama2024}
\nocite{yue2024mammoth2}
\nocite{wang2024backdooralign}
\nocite{hsu2024safe}
\nocite{hammoud-etal-2024-model}
\bibliography{anthology,custom}

\begin{thebibliography}{43}
\expandafter\ifx\csname natexlab\endcsname\relax\def\natexlab#1{#1}\fi

\bibitem[{Artetxe et~al.(2020)Artetxe, Ruder, and Yogatama}]{Artetxe_2020}
Mikel Artetxe, Sebastian Ruder, and Dani Yogatama. 2020.
\newblock \href {https://doi.org/10.18653/v1/2020.acl-main.421} {On the cross-lingual transferability of monolingual representations}.
\newblock In \emph{Proceedings of the 58th Annual Meeting of the Association for Computational Linguistics}. Association for Computational Linguistics.

\bibitem[{Bhardwaj et~al.(2024)Bhardwaj, Do, and Poria}]{bhardwaj-etal-2024-language}
Rishabh Bhardwaj, Duc~Anh Do, and Soujanya Poria. 2024.
\newblock \href {https://doi.org/10.18653/v1/2024.acl-long.762} {Language models are {H}omer simpson! safety re-alignment of fine-tuned language models through task arithmetic}.
\newblock In \emph{Proceedings of the 62nd Annual Meeting of the Association for Computational Linguistics (Volume 1: Long Papers)}, pages 14138--14149, Bangkok, Thailand. Association for Computational Linguistics.

\bibitem[{Bowen et~al.(2024)Bowen, Songning, Jiemin, Zhihao, Shiming, and Yutao}]{bowen2024taskvectorsselectivetask}
Tian Bowen, Lai Songning, Wu~Jiemin, Shuai Zhihao, Ge~Shiming, and Yue Yutao. 2024.
\newblock \href {http://arxiv.org/abs/2411.16139} {Beyond task vectors: Selective task arithmetic based on importance metrics}.

\bibitem[{Chen et~al.(2021)Chen, Tworek, Jun, Yuan, Pinto, Kaplan, Edwards, Burda, Joseph, Brockman et~al.}]{chen2021evaluating}
Mark Chen, Jerry Tworek, Heewoo Jun, Qiming Yuan, Henrique Ponde De~Oliveira Pinto, Jared Kaplan, Harri Edwards, Yuri Burda, Nicholas Joseph, Greg Brockman, et~al. 2021.
\newblock Evaluating large language models trained on code.
\newblock \emph{arXiv preprint arXiv:2107.03374}.

\bibitem[{Choi et~al.(2024)Choi, Kim, Lee, and Hong}]{choi2024}
Jiho Choi, Donggyun Kim, Chanhyuk Lee, and Seunghoon Hong. 2024.
\newblock Revisiting weight averaging for model merging.
\newblock \emph{arXiv preprint arXiv:2412.12153}.

\bibitem[{Cobbe et~al.(2021)Cobbe, Kosaraju, Bavarian, Chen, Jun, Kaiser, Plappert, Tworek, Hilton, Nakano, Hesse, and Schulman}]{Cobbe2021TrainingVT}
Karl Cobbe, Vineet Kosaraju, Mohammad Bavarian, Mark Chen, Heewoo Jun, Lukasz Kaiser, Matthias Plappert, Jerry Tworek, Jacob Hilton, Reiichiro Nakano, Christopher Hesse, and John Schulman. 2021.
\newblock \href {https://api.semanticscholar.org/CorpusID:239998651} {Training verifiers to solve math word problems}.
\newblock \emph{ArXiv}, abs/2110.14168.

\bibitem[{Conneau et~al.(2018)Conneau, Rinott, Lample, Williams, Bowman, Schwenk, and Stoyanov}]{conneau-etal-2018-xnli}
Alexis Conneau, Ruty Rinott, Guillaume Lample, Adina Williams, Samuel Bowman, Holger Schwenk, and Veselin Stoyanov. 2018.
\newblock \href {https://doi.org/10.18653/v1/D18-1269} {{XNLI}: Evaluating cross-lingual sentence representations}.
\newblock In \emph{Proceedings of the 2018 Conference on Empirical Methods in Natural Language Processing}, pages 2475--2485, Brussels, Belgium. Association for Computational Linguistics.

\bibitem[{Dai et~al.(2025)Dai, Hu, Shen, Zhang, Tian, and Ye}]{dai2025leveraging}
Rui Dai, Sile Hu, Xu~Shen, Yonggang Zhang, Xinmei Tian, and Jieping Ye. 2025.
\newblock Leveraging submodule linearity enhances task arithmetic performance in {LLM}s.
\newblock In \emph{The Thirteenth International Conference on Learning Representations (ICLR)}.

\bibitem[{Dodge et~al.(2020)Dodge, Ilharco, Schwartz, Farhadi, Hajishirzi, and Smith}]{dodge2020fine}
Jesse Dodge, Gabriel Ilharco, Roy Schwartz, Ali Farhadi, Hannaneh Hajishirzi, and Noah Smith. 2020.
\newblock Fine-tuning pretrained language models: Weight initializations, data orders, and early stopping.
\newblock \emph{arXiv preprint arXiv:2002.06305}.

\bibitem[{Du et~al.(2024)Du, Lee, Li, Jiang, Guo, Yu, Liu, Goh, Tang, He, and Zhang}]{guodong24neurips}
Guodong Du, Junlin Lee, Jing Li, Runhua Jiang, Yifei Guo, Shuyang Yu, Hanting Liu, Sim~Kuan Goh, Ho-Kin Tang, Daojing He, and Min Zhang. 2024.
\newblock Parameter competition balancing for model merging.
\newblock In \emph{The Thirty-eighth Annual Conference on Neural Information Processing Systems (NeurIPS)}.

\bibitem[{Gargiulo et~al.(2025)Gargiulo, Crisostomi, Bucarelli, Scardapane, Silvestri, and Rodolà}]{gargiulo2025tasksingularvectorsreducing}
Antonio~Andrea Gargiulo, Donato Crisostomi, Maria~Sofia Bucarelli, Simone Scardapane, Fabrizio Silvestri, and Emanuele Rodolà. 2025.
\newblock \href {http://arxiv.org/abs/2412.00081} {Task singular vectors: Reducing task interference in model merging}.

\bibitem[{Goddard et~al.(2024)Goddard, Siriwardhana, Ehghaghi, Meyers, Karpukhin, Benedict, McQuade, and Solawetz}]{goddard-etal-2024-arcees}
Charles Goddard, Shamane Siriwardhana, Malikeh Ehghaghi, Luke Meyers, Vladimir Karpukhin, Brian Benedict, Mark McQuade, and Jacob Solawetz. 2024.
\newblock \href {https://doi.org/10.18653/v1/2024.emnlp-industry.36} {Arcee{'}s {M}erge{K}it: A toolkit for merging large language models}.
\newblock In \emph{Proceedings of the 2024 Conference on Empirical Methods in Natural Language Processing: Industry Track}, pages 477--485, Miami, Florida, US. Association for Computational Linguistics.

\bibitem[{Grattafiori et~al.(2024)Grattafiori, Dubey, Jauhri, Pandey, Kadian, Al-Dahle, Letman, Mathur, Schelten, Vaughan, Yang, Fan, Goyal, Hartshorn, Yang, Mitra, Sravankumar, Korenev, Hinsvark, Rao, Zhang, Rodriguez, Gregerson, Spataru, Roziere, Biron, Tang, Chern, Caucheteux, Nayak, Bi, Marra, McConnell, Keller, Touret, Wu, Wong, Ferrer, Nikolaidis, Allonsius, Song, Pintz, Livshits, Wyatt, Esiobu, Choudhary, Mahajan, Garcia-Olano, Perino, Hupkes, Lakomkin, AlBadawy, Lobanova, Dinan, Smith, Radenovic, Guzmán, Zhang, Synnaeve, Lee, Anderson, Thattai, Nail, Mialon, Pang, Cucurell, Nguyen, Korevaar, Xu, Touvron, Zarov, Ibarra, Kloumann, Misra, Evtimov, Zhang, Copet, Lee, Geffert, Vranes, Park, Mahadeokar, Shah, van~der Linde, Billock, Hong, Lee, Fu, Chi, Huang, Liu, Wang, Yu, Bitton, Spisak, Park, Rocca, Johnstun, Saxe, Jia, Alwala, Prasad, Upasani, Plawiak, Li, Heafield, Stone, El-Arini, Iyer, Malik, Chiu, Bhalla, Lakhotia, Rantala-Yeary, van~der Maaten, Chen, Tan, Jenkins, Martin, Madaan, Malo, Blecher,
  Landzaat, de~Oliveira, Muzzi, Pasupuleti, Singh, Paluri, Kardas, Tsimpoukelli, Oldham, Rita, Pavlova, Kambadur, Lewis, Si, Singh, Hassan, Goyal, Torabi, Bashlykov, Bogoychev, Chatterji, Zhang, Duchenne, Çelebi, Alrassy, Zhang, Li, Vasic, Weng, Bhargava, Dubal, Krishnan, Koura, Xu, He, Dong, Srinivasan, Ganapathy, Calderer, Cabral, Stojnic, Raileanu, Maheswari, Girdhar, Patel, Sauvestre, Polidoro, Sumbaly, Taylor, Silva, Hou, Wang, Hosseini, Chennabasappa, Singh, Bell, Kim, Edunov, Nie, Narang, Raparthy, Shen, Wan, Bhosale, Zhang, Vandenhende, Batra, Whitman, Sootla, Collot, Gururangan, Borodinsky, Herman, Fowler, Sheasha, Georgiou, Scialom, Speckbacher, Mihaylov, Xiao, Karn, Goswami, Gupta, Ramanathan, Kerkez, Gonguet, Do, Vogeti, Albiero, Petrovic, Chu, Xiong, Fu, Meers, Martinet, Wang, Wang, Tan, Xia, Xie, Jia, Wang, Goldschlag, Gaur, Babaei, Wen, Song, Zhang, Li, Mao, Coudert, Yan, Chen, Papakipos, Singh, Srivastava, Jain, Kelsey, Shajnfeld, Gangidi, Victoria, Goldstand, Menon, Sharma, Boesenberg,
  Baevski, Feinstein, Kallet, Sangani, Teo, Yunus, Lupu, Alvarado, Caples, Gu, Ho, Poulton, Ryan, Ramchandani, Dong, Franco, Goyal, Saraf, Chowdhury, Gabriel, Bharambe, Eisenman, Yazdan, James, Maurer, Leonhardi, Huang, Loyd, Paola, Paranjape, Liu, Wu, Ni, Hancock, Wasti, Spence, Stojkovic, Gamido, Montalvo, Parker, Burton, Mejia, Liu, Wang, Kim, Zhou, Hu, Chu, Cai, Tindal, Feichtenhofer, Gao, Civin, Beaty, Kreymer, Li, Adkins, Xu, Testuggine, David, Parikh, Liskovich, Foss, Wang, Le, Holland, Dowling, Jamil, Montgomery, Presani, Hahn, Wood, Le, Brinkman, Arcaute, Dunbar, Smothers, Sun, Kreuk, Tian, Kokkinos, Ozgenel, Caggioni, Kanayet, Seide, Florez, Schwarz, Badeer, Swee, Halpern, Herman, Sizov, Guangyi, Zhang, Lakshminarayanan, Inan, Shojanazeri, Zou, Wang, Zha, Habeeb, Rudolph, Suk, Aspegren, Goldman, Zhan, Damlaj, Molybog, Tufanov, Leontiadis, Veliche, Gat, Weissman, Geboski, Kohli, Lam, Asher, Gaya, Marcus, Tang, Chan, Zhen, Reizenstein, Teboul, Zhong, Jin, Yang, Cummings, Carvill, Shepard, McPhie,
  Torres, Ginsburg, Wang, Wu, U, Saxena, Khandelwal, Zand, Matosich, Veeraraghavan, Michelena, Li, Jagadeesh, Huang, Chawla, Huang, Chen, Garg, A, Silva, Bell, Zhang, Guo, Yu, Moshkovich, Wehrstedt, Khabsa, Avalani, Bhatt, Mankus, Hasson, Lennie, Reso, Groshev, Naumov, Lathi, Keneally, Liu, Seltzer, Valko, Restrepo, Patel, Vyatskov, Samvelyan, Clark, Macey, Wang, Hermoso, Metanat, Rastegari, Bansal, Santhanam, Parks, White, Bawa, Singhal, Egebo, Usunier, Mehta, Laptev, Dong, Cheng, Chernoguz, Hart, Salpekar, Kalinli, Kent, Parekh, Saab, Balaji, Rittner, Bontrager, Roux, Dollar, Zvyagina, Ratanchandani, Yuvraj, Liang, Alao, Rodriguez, Ayub, Murthy, Nayani, Mitra, Parthasarathy, Li, Hogan, Battey, Wang, Howes, Rinott, Mehta, Siby, Bondu, Datta, Chugh, Hunt, Dhillon, Sidorov, Pan, Mahajan, Verma, Yamamoto, Ramaswamy, Lindsay, Lindsay, Feng, Lin, Zha, Patil, Shankar, Zhang, Zhang, Wang, Agarwal, Sajuyigbe, Chintala, Max, Chen, Kehoe, Satterfield, Govindaprasad, Gupta, Deng, Cho, Virk, Subramanian, Choudhury,
  Goldman, Remez, Glaser, Best, Koehler, Robinson, Li, Zhang, Matthews, Chou, Shaked, Vontimitta, Ajayi, Montanez, Mohan, Kumar, Mangla, Ionescu, Poenaru, Mihailescu, Ivanov, Li, Wang, Jiang, Bouaziz, Constable, Tang, Wu, Wang, Wu, Gao, Kleinman, Chen, Hu, Jia, Qi, Li, Zhang, Zhang, Adi, Nam, Yu, Wang, Zhao, Hao, Qian, Li, He, Rait, DeVito, Rosnbrick, Wen, Yang, Zhao, and Ma}]{grattafiori2024llama3herdmodels}
Aaron Grattafiori, Abhimanyu Dubey, Abhinav Jauhri, Abhinav Pandey, Abhishek Kadian, Ahmad Al-Dahle, Aiesha Letman, Akhil Mathur, Alan Schelten, Alex Vaughan, Amy Yang, Angela Fan, Anirudh Goyal, Anthony Hartshorn, Aobo Yang, Archi Mitra, Archie Sravankumar, Artem Korenev, Arthur Hinsvark, Arun Rao, Aston Zhang, Aurelien Rodriguez, Austen Gregerson, Ava Spataru, Baptiste Roziere, Bethany Biron, Binh Tang, Bobbie Chern, Charlotte Caucheteux, Chaya Nayak, Chloe Bi, Chris Marra, Chris McConnell, Christian Keller, Christophe Touret, Chunyang Wu, Corinne Wong, Cristian~Canton Ferrer, Cyrus Nikolaidis, Damien Allonsius, Daniel Song, Danielle Pintz, Danny Livshits, Danny Wyatt, David Esiobu, Dhruv Choudhary, Dhruv Mahajan, Diego Garcia-Olano, Diego Perino, Dieuwke Hupkes, Egor Lakomkin, Ehab AlBadawy, Elina Lobanova, Emily Dinan, Eric~Michael Smith, Filip Radenovic, Francisco Guzmán, Frank Zhang, Gabriel Synnaeve, Gabrielle Lee, Georgia~Lewis Anderson, Govind Thattai, Graeme Nail, Gregoire Mialon, Guan Pang,
  Guillem Cucurell, Hailey Nguyen, Hannah Korevaar, Hu~Xu, Hugo Touvron, Iliyan Zarov, Imanol~Arrieta Ibarra, Isabel Kloumann, Ishan Misra, Ivan Evtimov, Jack Zhang, Jade Copet, Jaewon Lee, Jan Geffert, Jana Vranes, Jason Park, Jay Mahadeokar, Jeet Shah, Jelmer van~der Linde, Jennifer Billock, Jenny Hong, Jenya Lee, Jeremy Fu, Jianfeng Chi, Jianyu Huang, Jiawen Liu, Jie Wang, Jiecao Yu, Joanna Bitton, Joe Spisak, Jongsoo Park, Joseph Rocca, Joshua Johnstun, Joshua Saxe, Junteng Jia, Kalyan~Vasuden Alwala, Karthik Prasad, Kartikeya Upasani, Kate Plawiak, Ke~Li, Kenneth Heafield, Kevin Stone, Khalid El-Arini, Krithika Iyer, Kshitiz Malik, Kuenley Chiu, Kunal Bhalla, Kushal Lakhotia, Lauren Rantala-Yeary, Laurens van~der Maaten, Lawrence Chen, Liang Tan, Liz Jenkins, Louis Martin, Lovish Madaan, Lubo Malo, Lukas Blecher, Lukas Landzaat, Luke de~Oliveira, Madeline Muzzi, Mahesh Pasupuleti, Mannat Singh, Manohar Paluri, Marcin Kardas, Maria Tsimpoukelli, Mathew Oldham, Mathieu Rita, Maya Pavlova, Melanie Kambadur,
  Mike Lewis, Min Si, Mitesh~Kumar Singh, Mona Hassan, Naman Goyal, Narjes Torabi, Nikolay Bashlykov, Nikolay Bogoychev, Niladri Chatterji, Ning Zhang, Olivier Duchenne, Onur Çelebi, Patrick Alrassy, Pengchuan Zhang, Pengwei Li, Petar Vasic, Peter Weng, Prajjwal Bhargava, Pratik Dubal, Praveen Krishnan, Punit~Singh Koura, Puxin Xu, Qing He, Qingxiao Dong, Ragavan Srinivasan, Raj Ganapathy, Ramon Calderer, Ricardo~Silveira Cabral, Robert Stojnic, Roberta Raileanu, Rohan Maheswari, Rohit Girdhar, Rohit Patel, Romain Sauvestre, Ronnie Polidoro, Roshan Sumbaly, Ross Taylor, Ruan Silva, Rui Hou, Rui Wang, Saghar Hosseini, Sahana Chennabasappa, Sanjay Singh, Sean Bell, Seohyun~Sonia Kim, Sergey Edunov, Shaoliang Nie, Sharan Narang, Sharath Raparthy, Sheng Shen, Shengye Wan, Shruti Bhosale, Shun Zhang, Simon Vandenhende, Soumya Batra, Spencer Whitman, Sten Sootla, Stephane Collot, Suchin Gururangan, Sydney Borodinsky, Tamar Herman, Tara Fowler, Tarek Sheasha, Thomas Georgiou, Thomas Scialom, Tobias Speckbacher,
  Todor Mihaylov, Tong Xiao, Ujjwal Karn, Vedanuj Goswami, Vibhor Gupta, Vignesh Ramanathan, Viktor Kerkez, Vincent Gonguet, Virginie Do, Vish Vogeti, Vítor Albiero, Vladan Petrovic, Weiwei Chu, Wenhan Xiong, Wenyin Fu, Whitney Meers, Xavier Martinet, Xiaodong Wang, Xiaofang Wang, Xiaoqing~Ellen Tan, Xide Xia, Xinfeng Xie, Xuchao Jia, Xuewei Wang, Yaelle Goldschlag, Yashesh Gaur, Yasmine Babaei, Yi~Wen, Yiwen Song, Yuchen Zhang, Yue Li, Yuning Mao, Zacharie~Delpierre Coudert, Zheng Yan, Zhengxing Chen, Zoe Papakipos, Aaditya Singh, Aayushi Srivastava, Abha Jain, Adam Kelsey, Adam Shajnfeld, Adithya Gangidi, Adolfo Victoria, Ahuva Goldstand, Ajay Menon, Ajay Sharma, Alex Boesenberg, Alexei Baevski, Allie Feinstein, Amanda Kallet, Amit Sangani, Amos Teo, Anam Yunus, Andrei Lupu, Andres Alvarado, Andrew Caples, Andrew Gu, Andrew Ho, Andrew Poulton, Andrew Ryan, Ankit Ramchandani, Annie Dong, Annie Franco, Anuj Goyal, Aparajita Saraf, Arkabandhu Chowdhury, Ashley Gabriel, Ashwin Bharambe, Assaf Eisenman, Azadeh
  Yazdan, Beau James, Ben Maurer, Benjamin Leonhardi, Bernie Huang, Beth Loyd, Beto~De Paola, Bhargavi Paranjape, Bing Liu, Bo~Wu, Boyu Ni, Braden Hancock, Bram Wasti, Brandon Spence, Brani Stojkovic, Brian Gamido, Britt Montalvo, Carl Parker, Carly Burton, Catalina Mejia, Ce~Liu, Changhan Wang, Changkyu Kim, Chao Zhou, Chester Hu, Ching-Hsiang Chu, Chris Cai, Chris Tindal, Christoph Feichtenhofer, Cynthia Gao, Damon Civin, Dana Beaty, Daniel Kreymer, Daniel Li, David Adkins, David Xu, Davide Testuggine, Delia David, Devi Parikh, Diana Liskovich, Didem Foss, Dingkang Wang, Duc Le, Dustin Holland, Edward Dowling, Eissa Jamil, Elaine Montgomery, Eleonora Presani, Emily Hahn, Emily Wood, Eric-Tuan Le, Erik Brinkman, Esteban Arcaute, Evan Dunbar, Evan Smothers, Fei Sun, Felix Kreuk, Feng Tian, Filippos Kokkinos, Firat Ozgenel, Francesco Caggioni, Frank Kanayet, Frank Seide, Gabriela~Medina Florez, Gabriella Schwarz, Gada Badeer, Georgia Swee, Gil Halpern, Grant Herman, Grigory Sizov, Guangyi, Zhang, Guna
  Lakshminarayanan, Hakan Inan, Hamid Shojanazeri, Han Zou, Hannah Wang, Hanwen Zha, Haroun Habeeb, Harrison Rudolph, Helen Suk, Henry Aspegren, Hunter Goldman, Hongyuan Zhan, Ibrahim Damlaj, Igor Molybog, Igor Tufanov, Ilias Leontiadis, Irina-Elena Veliche, Itai Gat, Jake Weissman, James Geboski, James Kohli, Janice Lam, Japhet Asher, Jean-Baptiste Gaya, Jeff Marcus, Jeff Tang, Jennifer Chan, Jenny Zhen, Jeremy Reizenstein, Jeremy Teboul, Jessica Zhong, Jian Jin, Jingyi Yang, Joe Cummings, Jon Carvill, Jon Shepard, Jonathan McPhie, Jonathan Torres, Josh Ginsburg, Junjie Wang, Kai Wu, Kam~Hou U, Karan Saxena, Kartikay Khandelwal, Katayoun Zand, Kathy Matosich, Kaushik Veeraraghavan, Kelly Michelena, Keqian Li, Kiran Jagadeesh, Kun Huang, Kunal Chawla, Kyle Huang, Lailin Chen, Lakshya Garg, Lavender A, Leandro Silva, Lee Bell, Lei Zhang, Liangpeng Guo, Licheng Yu, Liron Moshkovich, Luca Wehrstedt, Madian Khabsa, Manav Avalani, Manish Bhatt, Martynas Mankus, Matan Hasson, Matthew Lennie, Matthias Reso, Maxim
  Groshev, Maxim Naumov, Maya Lathi, Meghan Keneally, Miao Liu, Michael~L. Seltzer, Michal Valko, Michelle Restrepo, Mihir Patel, Mik Vyatskov, Mikayel Samvelyan, Mike Clark, Mike Macey, Mike Wang, Miquel~Jubert Hermoso, Mo~Metanat, Mohammad Rastegari, Munish Bansal, Nandhini Santhanam, Natascha Parks, Natasha White, Navyata Bawa, Nayan Singhal, Nick Egebo, Nicolas Usunier, Nikhil Mehta, Nikolay~Pavlovich Laptev, Ning Dong, Norman Cheng, Oleg Chernoguz, Olivia Hart, Omkar Salpekar, Ozlem Kalinli, Parkin Kent, Parth Parekh, Paul Saab, Pavan Balaji, Pedro Rittner, Philip Bontrager, Pierre Roux, Piotr Dollar, Polina Zvyagina, Prashant Ratanchandani, Pritish Yuvraj, Qian Liang, Rachad Alao, Rachel Rodriguez, Rafi Ayub, Raghotham Murthy, Raghu Nayani, Rahul Mitra, Rangaprabhu Parthasarathy, Raymond Li, Rebekkah Hogan, Robin Battey, Rocky Wang, Russ Howes, Ruty Rinott, Sachin Mehta, Sachin Siby, Sai~Jayesh Bondu, Samyak Datta, Sara Chugh, Sara Hunt, Sargun Dhillon, Sasha Sidorov, Satadru Pan, Saurabh Mahajan,
  Saurabh Verma, Seiji Yamamoto, Sharadh Ramaswamy, Shaun Lindsay, Shaun Lindsay, Sheng Feng, Shenghao Lin, Shengxin~Cindy Zha, Shishir Patil, Shiva Shankar, Shuqiang Zhang, Shuqiang Zhang, Sinong Wang, Sneha Agarwal, Soji Sajuyigbe, Soumith Chintala, Stephanie Max, Stephen Chen, Steve Kehoe, Steve Satterfield, Sudarshan Govindaprasad, Sumit Gupta, Summer Deng, Sungmin Cho, Sunny Virk, Suraj Subramanian, Sy~Choudhury, Sydney Goldman, Tal Remez, Tamar Glaser, Tamara Best, Thilo Koehler, Thomas Robinson, Tianhe Li, Tianjun Zhang, Tim Matthews, Timothy Chou, Tzook Shaked, Varun Vontimitta, Victoria Ajayi, Victoria Montanez, Vijai Mohan, Vinay~Satish Kumar, Vishal Mangla, Vlad Ionescu, Vlad Poenaru, Vlad~Tiberiu Mihailescu, Vladimir Ivanov, Wei Li, Wenchen Wang, Wenwen Jiang, Wes Bouaziz, Will Constable, Xiaocheng Tang, Xiaojian Wu, Xiaolan Wang, Xilun Wu, Xinbo Gao, Yaniv Kleinman, Yanjun Chen, Ye~Hu, Ye~Jia, Ye~Qi, Yenda Li, Yilin Zhang, Ying Zhang, Yossi Adi, Youngjin Nam, Yu, Wang, Yu~Zhao, Yuchen Hao, Yundi
  Qian, Yunlu Li, Yuzi He, Zach Rait, Zachary DeVito, Zef Rosnbrick, Zhaoduo Wen, Zhenyu Yang, Zhiwei Zhao, and Zhiyu Ma. 2024.
\newblock \href {http://arxiv.org/abs/2407.21783} {The llama 3 herd of models}.

\bibitem[{Hammoud et~al.(2024)Hammoud, Michieli, Pizzati, Torr, Bibi, Ghanem, and Ozay}]{hammoud-etal-2024-model}
Hasan Abed Al~Kader Hammoud, Umberto Michieli, Fabio Pizzati, Philip Torr, Adel Bibi, Bernard Ghanem, and Mete Ozay. 2024.
\newblock \href {https://doi.org/10.18653/v1/2024.findings-emnlp.762} {Model merging and safety alignment: One bad model spoils the bunch}.
\newblock In \emph{Findings of the Association for Computational Linguistics: EMNLP 2024}, pages 13033--13046, Miami, Florida, USA. Association for Computational Linguistics.

\bibitem[{Hazra et~al.(2024)Hazra, Layek, Banerjee, and Poria}]{hazra-etal-2024-safety}
Rima Hazra, Sayan Layek, Somnath Banerjee, and Soujanya Poria. 2024.
\newblock \href {https://doi.org/10.18653/v1/2024.emnlp-main.1212} {Safety arithmetic: A framework for test-time safety alignment of language models by steering parameters and activations}.
\newblock In \emph{Proceedings of the 2024 Conference on Empirical Methods in Natural Language Processing}, pages 21759--21776, Miami, Florida, USA. Association for Computational Linguistics.

\bibitem[{Hirakawa et~al.(2024)Hirakawa, Horie, Nakamura, Oba, Passaglia, and Sasaki}]{elyzallama2024}
Masato Hirakawa, Shintaro Horie, Tomoaki Nakamura, Daisuke Oba, Sam Passaglia, and Akira Sasaki. 2024.
\newblock \href {https://huggingface.co/elyza/Llama-3-ELYZA-JP-8B} {elyza/llama-3-elyza-jp-8b}.

\bibitem[{Hsu et~al.(2024)Hsu, Tsai, Lin, Chen, Yu, and Huang}]{hsu2024safe}
Chia-Yi Hsu, Yu-Lin Tsai, Chih-Hsun Lin, Pin-Yu Chen, Chia-Mu Yu, and Chun-Ying Huang. 2024.
\newblock \href {https://openreview.net/forum?id=HcifdQZFZV} {Safe lo{RA}: The silver lining of reducing safety risks when finetuning large language models}.
\newblock In \emph{The Thirty-eighth Annual Conference on Neural Information Processing Systems}.

\bibitem[{Huang et~al.(2024)Huang, Li, Hsu, Chen, Lin, Hsiao, Tsai, and Lee}]{huang-etal-2024-chat}
Shih-Cheng Huang, Pin-Zu Li, Yu-chi Hsu, Kuang-Ming Chen, Yu~Tung Lin, Shih-Kai Hsiao, Richard Tsai, and Hung-yi Lee. 2024.
\newblock \href {https://doi.org/10.18653/v1/2024.acl-long.590} {Chat vector: A simple approach to equip {LLM}s with instruction following and model alignment in new languages}.
\newblock In \emph{Proceedings of the 62nd Annual Meeting of the Association for Computational Linguistics (Volume 1: Long Papers)}, pages 10943--10959, Bangkok, Thailand. Association for Computational Linguistics.

\bibitem[{Ilharco et~al.(2023)Ilharco, Ribeiro, Wortsman, Gururangan, Schmidt, Hajishirzi, and Farhadi}]{ilharco2023}
Gabriel Ilharco, Marco~Tulio Ribeiro, Mitchell Wortsman, Suchin Gururangan, Ludwig Schmidt, Hannaneh Hajishirzi, and Ali Farhadi. 2023.
\newblock Editing models with task arithmetic.
\newblock In \emph{Proceedings of the 11th International Conference on Learning Representations (ICLR)}.

\bibitem[{Jin et~al.(2023)Jin, Ren, Preotiuc-Pietro, and Cheng}]{jin2023dataless}
Xisen Jin, Xiang Ren, Daniel Preotiuc-Pietro, and Pengxiang Cheng. 2023.
\newblock \href {https://openreview.net/forum?id=FCnohuR6AnM} {Dataless knowledge fusion by merging weights of language models}.
\newblock In \emph{The Eleventh International Conference on Learning Representations}.

\bibitem[{Kurihara et~al.(2022)Kurihara, Kawahara, and Shibata}]{kurihara-etal-2022-jglue}
Kentaro Kurihara, Daisuke Kawahara, and Tomohide Shibata. 2022.
\newblock \href {https://aclanthology.org/2022.lrec-1.317} {{JGLUE}: {J}apanese general language understanding evaluation}.
\newblock In \emph{Proceedings of the Thirteenth Language Resources and Evaluation Conference}, pages 2957--2966, Marseille, France. European Language Resources Association.

\bibitem[{Lai et~al.(2025)Lai, Tang, Pan, Dong, Liu, Chen, Shen, Li, and Chu}]{lai2025mediatormemoryefficientllmmerging}
Kunfeng Lai, Zhenheng Tang, Xinglin Pan, Peijie Dong, Xiang Liu, Haolan Chen, Li~Shen, Bo~Li, and Xiaowen Chu. 2025.
\newblock \href {http://arxiv.org/abs/2502.04411} {Mediator: Memory-efficient llm merging with less parameter conflicts and uncertainty based routing}.

\bibitem[{Lai et~al.(2023)Lai, Nguyen, Ngo, Nguyen, Dernoncourt, Rossi, and Nguyen}]{lai-etal-2023-okapi}
Viet Lai, Chien Nguyen, Nghia Ngo, Thuat Nguyen, Franck Dernoncourt, Ryan Rossi, and Thien Nguyen. 2023.
\newblock \href {https://doi.org/10.18653/v1/2023.emnlp-demo.28} {Okapi: Instruction-tuned large language models in multiple languages with reinforcement learning from human feedback}.
\newblock In \emph{Proceedings of the 2023 Conference on Empirical Methods in Natural Language Processing: System Demonstrations}, pages 318--327, Singapore. Association for Computational Linguistics.

\bibitem[{Li et~al.(2025)Li, Yao, Zhang, and Li}]{li2025safety}
Shen Li, Liuyi Yao, Lan Zhang, and Yaliang Li. 2025.
\newblock \href {https://openreview.net/forum?id=kUH1yPMAn7} {Safety layers in aligned large language models: The key to {LLM} security}.
\newblock In \emph{The Thirteenth International Conference on Learning Representations}.

\bibitem[{Lu et~al.(2024)Lu, Fan, Wei, Qu, Chen, and Cheng}]{lu2024}
Zhenyi Lu, Chenghao Fan, Wei Wei, Xiaoye Qu, Dangyang Chen, and Yu~Cheng. 2024.
\newblock Twin-merging: Dynamic integration of modular expertise in model merging.
\newblock In \emph{Advances in Neural Information Processing Systems (NeurIPS)}.

\bibitem[{Matena and Raffel(2022)}]{matena2022}
Michael~S. Matena and Colin~A. Raffel. 2022.
\newblock Merging models with fisher-weighted averaging.
\newblock In \emph{Advances in Neural Information Processing Systems (NeurIPS)}.

\bibitem[{Merity et~al.(2017)Merity, Xiong, Bradbury, and Socher}]{merity2017pointer}
Stephen Merity, Caiming Xiong, James Bradbury, and Richard Socher. 2017.
\newblock \href {https://openreview.net/forum?id=Byj72udxe} {Pointer sentinel mixture models}.
\newblock In \emph{International Conference on Learning Representations}.

\bibitem[{OpenAI et~al.(2024)OpenAI, Achiam, Adler, Agarwal, Ahmad, Akkaya, Aleman, Almeida, Altenschmidt, Altman, Anadkat, Avila, Babuschkin, Balaji, Balcom, Baltescu, Bao, Bavarian, Belgum, Bello, Berdine, Bernadett-Shapiro, Berner, Bogdonoff, Boiko, Boyd, Brakman, Brockman, Brooks, Brundage, Button, Cai, Campbell, Cann, Carey, Carlson, Carmichael, Chan, Chang, Chantzis, Chen, Chen, Chen, Chen, Chen, Chess, Cho, Chu, Chung, Cummings, Currier, Dai, Decareaux, Degry, Deutsch, Deville, Dhar, Dohan, Dowling, Dunning, Ecoffet, Eleti, Eloundou, Farhi, Fedus, Felix, Fishman, Forte, Fulford, Gao, Georges, Gibson, Goel, Gogineni, Goh, Gontijo-Lopes, Gordon, Grafstein, Gray, Greene, Gross, Gu, Guo, Hallacy, Han, Harris, He, Heaton, Heidecke, Hesse, Hickey, Hickey, Hoeschele, Houghton, Hsu, Hu, Hu, Huizinga, Jain, Jain, Jang, Jiang, Jiang, Jin, Jin, Jomoto, Jonn, Jun, Kaftan, Łukasz Kaiser, Kamali, Kanitscheider, Keskar, Khan, Kilpatrick, Kim, Kim, Kim, Kirchner, Kiros, Knight, Kokotajlo, Łukasz Kondraciuk,
  Kondrich, Konstantinidis, Kosic, Krueger, Kuo, Lampe, Lan, Lee, Leike, Leung, Levy, Li, Lim, Lin, Lin, Litwin, Lopez, Lowe, Lue, Makanju, Malfacini, Manning, Markov, Markovski, Martin, Mayer, Mayne, McGrew, McKinney, McLeavey, McMillan, McNeil, Medina, Mehta, Menick, Metz, Mishchenko, Mishkin, Monaco, Morikawa, Mossing, Mu, Murati, Murk, Mély, Nair, Nakano, Nayak, Neelakantan, Ngo, Noh, Ouyang, O'Keefe, Pachocki, Paino, Palermo, Pantuliano, Parascandolo, Parish, Parparita, Passos, Pavlov, Peng, Perelman, de~Avila Belbute~Peres, Petrov, de~Oliveira~Pinto, Michael, Pokorny, Pokrass, Pong, Powell, Power, Power, Proehl, Puri, Radford, Rae, Ramesh, Raymond, Real, Rimbach, Ross, Rotsted, Roussez, Ryder, Saltarelli, Sanders, Santurkar, Sastry, Schmidt, Schnurr, Schulman, Selsam, Sheppard, Sherbakov, Shieh, Shoker, Shyam, Sidor, Sigler, Simens, Sitkin, Slama, Sohl, Sokolowsky, Song, Staudacher, Such, Summers, Sutskever, Tang, Tezak, Thompson, Tillet, Tootoonchian, Tseng, Tuggle, Turley, Tworek, Uribe, Vallone,
  Vijayvergiya, Voss, Wainwright, Wang, Wang, Wang, Ward, Wei, Weinmann, Welihinda, Welinder, Weng, Weng, Wiethoff, Willner, Winter, Wolrich, Wong, Workman, Wu, Wu, Wu, Xiao, Xu, Yoo, Yu, Yuan, Zaremba, Zellers, Zhang, Zhang, Zhao, Zheng, Zhuang, Zhuk, and Zoph}]{openai2024gpt4technicalreport}
OpenAI, Josh Achiam, Steven Adler, Sandhini Agarwal, Lama Ahmad, Ilge Akkaya, Florencia~Leoni Aleman, Diogo Almeida, Janko Altenschmidt, Sam Altman, Shyamal Anadkat, Red Avila, Igor Babuschkin, Suchir Balaji, Valerie Balcom, Paul Baltescu, Haiming Bao, Mohammad Bavarian, Jeff Belgum, Irwan Bello, Jake Berdine, Gabriel Bernadett-Shapiro, Christopher Berner, Lenny Bogdonoff, Oleg Boiko, Madelaine Boyd, Anna-Luisa Brakman, Greg Brockman, Tim Brooks, Miles Brundage, Kevin Button, Trevor Cai, Rosie Campbell, Andrew Cann, Brittany Carey, Chelsea Carlson, Rory Carmichael, Brooke Chan, Che Chang, Fotis Chantzis, Derek Chen, Sully Chen, Ruby Chen, Jason Chen, Mark Chen, Ben Chess, Chester Cho, Casey Chu, Hyung~Won Chung, Dave Cummings, Jeremiah Currier, Yunxing Dai, Cory Decareaux, Thomas Degry, Noah Deutsch, Damien Deville, Arka Dhar, David Dohan, Steve Dowling, Sheila Dunning, Adrien Ecoffet, Atty Eleti, Tyna Eloundou, David Farhi, Liam Fedus, Niko Felix, Simón~Posada Fishman, Juston Forte, Isabella Fulford, Leo
  Gao, Elie Georges, Christian Gibson, Vik Goel, Tarun Gogineni, Gabriel Goh, Rapha Gontijo-Lopes, Jonathan Gordon, Morgan Grafstein, Scott Gray, Ryan Greene, Joshua Gross, Shixiang~Shane Gu, Yufei Guo, Chris Hallacy, Jesse Han, Jeff Harris, Yuchen He, Mike Heaton, Johannes Heidecke, Chris Hesse, Alan Hickey, Wade Hickey, Peter Hoeschele, Brandon Houghton, Kenny Hsu, Shengli Hu, Xin Hu, Joost Huizinga, Shantanu Jain, Shawn Jain, Joanne Jang, Angela Jiang, Roger Jiang, Haozhun Jin, Denny Jin, Shino Jomoto, Billie Jonn, Heewoo Jun, Tomer Kaftan, Łukasz Kaiser, Ali Kamali, Ingmar Kanitscheider, Nitish~Shirish Keskar, Tabarak Khan, Logan Kilpatrick, Jong~Wook Kim, Christina Kim, Yongjik Kim, Jan~Hendrik Kirchner, Jamie Kiros, Matt Knight, Daniel Kokotajlo, Łukasz Kondraciuk, Andrew Kondrich, Aris Konstantinidis, Kyle Kosic, Gretchen Krueger, Vishal Kuo, Michael Lampe, Ikai Lan, Teddy Lee, Jan Leike, Jade Leung, Daniel Levy, Chak~Ming Li, Rachel Lim, Molly Lin, Stephanie Lin, Mateusz Litwin, Theresa Lopez, Ryan
  Lowe, Patricia Lue, Anna Makanju, Kim Malfacini, Sam Manning, Todor Markov, Yaniv Markovski, Bianca Martin, Katie Mayer, Andrew Mayne, Bob McGrew, Scott~Mayer McKinney, Christine McLeavey, Paul McMillan, Jake McNeil, David Medina, Aalok Mehta, Jacob Menick, Luke Metz, Andrey Mishchenko, Pamela Mishkin, Vinnie Monaco, Evan Morikawa, Daniel Mossing, Tong Mu, Mira Murati, Oleg Murk, David Mély, Ashvin Nair, Reiichiro Nakano, Rajeev Nayak, Arvind Neelakantan, Richard Ngo, Hyeonwoo Noh, Long Ouyang, Cullen O'Keefe, Jakub Pachocki, Alex Paino, Joe Palermo, Ashley Pantuliano, Giambattista Parascandolo, Joel Parish, Emy Parparita, Alex Passos, Mikhail Pavlov, Andrew Peng, Adam Perelman, Filipe de~Avila Belbute~Peres, Michael Petrov, Henrique~Ponde de~Oliveira~Pinto, Michael, Pokorny, Michelle Pokrass, Vitchyr~H. Pong, Tolly Powell, Alethea Power, Boris Power, Elizabeth Proehl, Raul Puri, Alec Radford, Jack Rae, Aditya Ramesh, Cameron Raymond, Francis Real, Kendra Rimbach, Carl Ross, Bob Rotsted, Henri Roussez,
  Nick Ryder, Mario Saltarelli, Ted Sanders, Shibani Santurkar, Girish Sastry, Heather Schmidt, David Schnurr, John Schulman, Daniel Selsam, Kyla Sheppard, Toki Sherbakov, Jessica Shieh, Sarah Shoker, Pranav Shyam, Szymon Sidor, Eric Sigler, Maddie Simens, Jordan Sitkin, Katarina Slama, Ian Sohl, Benjamin Sokolowsky, Yang Song, Natalie Staudacher, Felipe~Petroski Such, Natalie Summers, Ilya Sutskever, Jie Tang, Nikolas Tezak, Madeleine~B. Thompson, Phil Tillet, Amin Tootoonchian, Elizabeth Tseng, Preston Tuggle, Nick Turley, Jerry Tworek, Juan Felipe~Cerón Uribe, Andrea Vallone, Arun Vijayvergiya, Chelsea Voss, Carroll Wainwright, Justin~Jay Wang, Alvin Wang, Ben Wang, Jonathan Ward, Jason Wei, CJ~Weinmann, Akila Welihinda, Peter Welinder, Jiayi Weng, Lilian Weng, Matt Wiethoff, Dave Willner, Clemens Winter, Samuel Wolrich, Hannah Wong, Lauren Workman, Sherwin Wu, Jeff Wu, Michael Wu, Kai Xiao, Tao Xu, Sarah Yoo, Kevin Yu, Qiming Yuan, Wojciech Zaremba, Rowan Zellers, Chong Zhang, Marvin Zhang, Shengjia
  Zhao, Tianhao Zheng, Juntang Zhuang, William Zhuk, and Barret Zoph. 2024.
\newblock \href {http://arxiv.org/abs/2303.08774} {Gpt-4 technical report}.

\bibitem[{Pipatanakul et~al.(2023)Pipatanakul, Jirabovonvisut, Manakul, Sripaisarnmongkol, Patomwong, Chokchainant, and Tharnpipitchai}]{pipatanakul2023typhoon}
Kunat Pipatanakul, Phatrasek Jirabovonvisut, Potsawee Manakul, Sittipong Sripaisarnmongkol, Ruangsak Patomwong, Pathomporn Chokchainant, and Kasima Tharnpipitchai. 2023.
\newblock \href {https://arxiv.org/abs/2312.13951} {Typhoon: Thai large language models}.
\newblock \emph{arXiv preprint arXiv:2312.13951}.

\bibitem[{Qi et~al.(2024)Qi, Zeng, Xie, Chen, Jia, Mittal, and Henderson}]{qi2024finetuning}
Xiangyu Qi, Yi~Zeng, Tinghao Xie, Pin-Yu Chen, Ruoxi Jia, Prateek Mittal, and Peter Henderson. 2024.
\newblock \href {https://openreview.net/forum?id=hTEGyKf0dZ} {Fine-tuning aligned language models compromises safety, even when users do not intend to!}
\newblock In \emph{The Twelfth International Conference on Learning Representations}.

\bibitem[{Riviere et~al.(2024)Riviere, Pathak, Sessa, Hardin, Bhupatiraju, Hussenot, Mesnard, Shahriari, Ramé, Ferret, Liu, Tafti, Friesen, Casbon, Ramos, Kumar, Lan, Jerome, Tsitsulin, Vieillard, Stanczyk, Girgin, Momchev, Hoffman, Thakoor, Grill, Neyshabur, Bachem, Walton, Severyn, Parrish, Ahmad, Hutchison, Abdagic, Carl, Shen, Brock, Coenen, Laforge, Paterson, Bastian, Piot, Wu, Royal, Chen, Kumar, Perry, Welty, Choquette-Choo, Sinopalnikov, Weinberger, Vijaykumar, Rogozińska, Herbison, Bandy, Wang, Noland, Moreira, Senter, Eltyshev, Visin, Rasskin, Wei, Cameron, Martins, Hashemi, Klimczak-Plucińska, Batra, Dhand, Nardini, Mein, Zhou, Svensson, Stanway, Chan, Zhou, Carrasqueira, Iljazi, Becker, Fernandez, van Amersfoort, Gordon, Lipschultz, Newlan, yeong Ji, Mohamed, Badola, Black, Millican, McDonell, Nguyen, Sodhia, Greene, Sjoesund, Usui, Sifre, Heuermann, Lago, McNealus, Soares, Kilpatrick, Dixon, Martins, Reid, Singh, Iverson, Görner, Velloso, Wirth, Davidow, Miller, Rahtz, Watson, Risdal, Kazemi,
  Moynihan, Zhang, Kahng, Park, Rahman, Khatwani, Dao, Bardoliwalla, Devanathan, Dumai, Chauhan, Wahltinez, Botarda, Barnes, Barham, Michel, Jin, Georgiev, Culliton, Kuppala, Comanescu, Merhej, Jana, Rokni, Agarwal, Mullins, Saadat, Carthy, Cogan, Perrin, Arnold, Krause, Dai, Garg, Sheth, Ronstrom, Chan, Jordan, Yu, Eccles, Hennigan, Kocisky, Doshi, Jain, Yadav, Meshram, Dharmadhikari, Barkley, Wei, Ye, Han, Kwon, Xu, Shen, Gong, Wei, Cotruta, Kirk, Rao, Giang, Peran, Warkentin, Collins, Barral, Ghahramani, Hadsell, Sculley, Banks, Dragan, Petrov, Vinyals, Dean, Hassabis, Kavukcuoglu, Farabet, Buchatskaya, Borgeaud, Fiedel, Joulin, Kenealy, Dadashi, and Andreev}]{gemmateam2024gemma2improvingopen}
Morgane Riviere, Shreya Pathak, Pier~Giuseppe Sessa, Cassidy Hardin, Surya Bhupatiraju, Léonard Hussenot, Thomas Mesnard, Bobak Shahriari, Alexandre Ramé, Johan Ferret, Peter Liu, Pouya Tafti, Abe Friesen, Michelle Casbon, Sabela Ramos, Ravin Kumar, Charline~Le Lan, Sammy Jerome, Anton Tsitsulin, Nino Vieillard, Piotr Stanczyk, Sertan Girgin, Nikola Momchev, Matt Hoffman, Shantanu Thakoor, Jean-Bastien Grill, Behnam Neyshabur, Olivier Bachem, Alanna Walton, Aliaksei Severyn, Alicia Parrish, Aliya Ahmad, Allen Hutchison, Alvin Abdagic, Amanda Carl, Amy Shen, Andy Brock, Andy Coenen, Anthony Laforge, Antonia Paterson, Ben Bastian, Bilal Piot, Bo~Wu, Brandon Royal, Charlie Chen, Chintu Kumar, Chris Perry, Chris Welty, Christopher~A. Choquette-Choo, Danila Sinopalnikov, David Weinberger, Dimple Vijaykumar, Dominika Rogozińska, Dustin Herbison, Elisa Bandy, Emma Wang, Eric Noland, Erica Moreira, Evan Senter, Evgenii Eltyshev, Francesco Visin, Gabriel Rasskin, Gary Wei, Glenn Cameron, Gus Martins, Hadi Hashemi,
  Hanna Klimczak-Plucińska, Harleen Batra, Harsh Dhand, Ivan Nardini, Jacinda Mein, Jack Zhou, James Svensson, Jeff Stanway, Jetha Chan, Jin~Peng Zhou, Joana Carrasqueira, Joana Iljazi, Jocelyn Becker, Joe Fernandez, Joost van Amersfoort, Josh Gordon, Josh Lipschultz, Josh Newlan, Ju~yeong Ji, Kareem Mohamed, Kartikeya Badola, Kat Black, Katie Millican, Keelin McDonell, Kelvin Nguyen, Kiranbir Sodhia, Kish Greene, Lars~Lowe Sjoesund, Lauren Usui, Laurent Sifre, Lena Heuermann, Leticia Lago, Lilly McNealus, Livio~Baldini Soares, Logan Kilpatrick, Lucas Dixon, Luciano Martins, Machel Reid, Manvinder Singh, Mark Iverson, Martin Görner, Mat Velloso, Mateo Wirth, Matt Davidow, Matt Miller, Matthew Rahtz, Matthew Watson, Meg Risdal, Mehran Kazemi, Michael Moynihan, Ming Zhang, Minsuk Kahng, Minwoo Park, Mofi Rahman, Mohit Khatwani, Natalie Dao, Nenshad Bardoliwalla, Nesh Devanathan, Neta Dumai, Nilay Chauhan, Oscar Wahltinez, Pankil Botarda, Parker Barnes, Paul Barham, Paul Michel, Pengchong Jin, Petko Georgiev,
  Phil Culliton, Pradeep Kuppala, Ramona Comanescu, Ramona Merhej, Reena Jana, Reza~Ardeshir Rokni, Rishabh Agarwal, Ryan Mullins, Samaneh Saadat, Sara~Mc Carthy, Sarah Cogan, Sarah Perrin, Sébastien M.~R. Arnold, Sebastian Krause, Shengyang Dai, Shruti Garg, Shruti Sheth, Sue Ronstrom, Susan Chan, Timothy Jordan, Ting Yu, Tom Eccles, Tom Hennigan, Tomas Kocisky, Tulsee Doshi, Vihan Jain, Vikas Yadav, Vilobh Meshram, Vishal Dharmadhikari, Warren Barkley, Wei Wei, Wenming Ye, Woohyun Han, Woosuk Kwon, Xiang Xu, Zhe Shen, Zhitao Gong, Zichuan Wei, Victor Cotruta, Phoebe Kirk, Anand Rao, Minh Giang, Ludovic Peran, Tris Warkentin, Eli Collins, Joelle Barral, Zoubin Ghahramani, Raia Hadsell, D.~Sculley, Jeanine Banks, Anca Dragan, Slav Petrov, Oriol Vinyals, Jeff Dean, Demis Hassabis, Koray Kavukcuoglu, Clement Farabet, Elena Buchatskaya, Sebastian Borgeaud, Noah Fiedel, Armand Joulin, Kathleen Kenealy, Robert Dadashi, and Alek Andreev. 2024.
\newblock \href {http://arxiv.org/abs/2408.00118} {Gemma 2: Improving open language models at a practical size}.

\bibitem[{Suzuki et~al.(2020)Suzuki, Suzuki, Matsuda, Nishida, and Inoue}]{suzuki2020jaqket}
Masatoshi Suzuki, Jun Suzuki, Koji Matsuda, Kyosuke Nishida, and Naoya Inoue. 2020.
\newblock {JAQKET}: Construction of a japanese {QA} dataset on the subject of quizzes.
\newblock In \emph{Proceedings of the Annual Meeting of the Association for Natural Language Processing}, volume~26, pages 237--240.

\bibitem[{Tam et~al.(2024)Tam, Pai, Lee, Shuai, Chen, Chu, and Cheng}]{tam2024tmmlu}
Zhi~Rui Tam, Ya~Ting Pai, Yen-Wei Lee, Hong-Han Shuai, Jun-Da Chen, Wei~Min Chu, and Sega Cheng. 2024.
\newblock \href {https://openreview.net/forum?id=95TayIeqJ4} {{TMMLU}+: An improved traditional chinese evaluation suite for foundation models}.
\newblock In \emph{First Conference on Language Modeling}.

\bibitem[{Wang et~al.(2024)Wang, Li, Li, Qi, Hu, Li, McDaniel, Chen, Li, and Xiao}]{wang2024backdooralign}
Jiongxiao Wang, Jiazhao Li, Yiquan Li, Xiangyu Qi, Junjie Hu, Yixuan Li, Patrick McDaniel, Muhao Chen, Bo~Li, and Chaowei Xiao. 2024.
\newblock \href {https://openreview.net/forum?id=1PcJ5Evta7} {Backdooralign: Mitigating fine-tuning based jailbreak attack with backdoor enhanced safety alignment}.
\newblock In \emph{The Thirty-eighth Annual Conference on Neural Information Processing Systems}.

\bibitem[{Wortsman et~al.(2022)Wortsman, Ilharco, Gadre, Roelofs, Gontijo-Lopes, Morcos, Namkoong, Farhadi, Carmon, Kornblith, and Schmidt}]{wortsman2022}
Mitchell Wortsman, Gabriel Ilharco, Samir~Ya Gadre, Rebecca Roelofs, Raphael Gontijo-Lopes, Ari~S. Morcos, Hongseok Namkoong, Ali Farhadi, Yair Carmon, Simon Kornblith, and Ludwig Schmidt. 2022.
\newblock Model soups: averaging weights of multiple fine-tuned models improves accuracy without increasing inference time.
\newblock In \emph{Proceedings of the 39th International Conference on Machine Learning}, pages 23965--23998. PMLR.

\bibitem[{Yadav et~al.(2023)Yadav, Tam, Choshen, Raffel, and Bansal}]{yadav2023}
Prateek Yadav, Derek Tam, Leshem Choshen, Colin Raffel, and Mohit Bansal. 2023.
\newblock Ties-merging: Resolving interference when merging models.
\newblock In \emph{Advances in Neural Information Processing Systems (NeurIPS)}.

\bibitem[{Yang et~al.(2024{\natexlab{a}})Yang, Shen, Wang, Guo, Chen, Wang, and Tao}]{yang2024representation}
Enneng Yang, Li~Shen, Zhenyi Wang, Guibing Guo, Xiaojun Chen, Xingwei Wang, and Dacheng Tao. 2024{\natexlab{a}}.
\newblock \href {https://openreview.net/forum?id=Sbl2keQEML} {Representation surgery for multi-task model merging}.
\newblock In \emph{Forty-first International Conference on Machine Learning}.

\bibitem[{Yang et~al.(2024{\natexlab{b}})Yang, Wang, Shen, Liu, Guo, Wang, and Tao}]{yang2024adamerging}
Enneng Yang, Zhenyi Wang, Li~Shen, Shiwei Liu, Guibing Guo, Xingwei Wang, and Dacheng Tao. 2024{\natexlab{b}}.
\newblock \href {https://openreview.net/forum?id=nZP6NgD3QY} {Adamerging: Adaptive model merging for multi-task learning}.
\newblock In \emph{The Twelfth International Conference on Learning Representations}.

\bibitem[{Yu et~al.(2024)Yu, Yu, Yu, Huang, and Li}]{yu2024}
Le~Yu, Bowen Yu, Haiyang Yu, Fei Huang, and Yongbin Li. 2024.
\newblock Language models are super mario: Absorbing abilities from homologous models as a free lunch.
\newblock In \emph{Proceedings of the 41st International Conference on Machine Learning (ICML)}.

\bibitem[{Yue et~al.(2024)Yue, Zheng, Zhang, and Chen}]{yue2024mammoth2}
Xiang Yue, Tuney Zheng, Ge~Zhang, and Wenhu Chen. 2024.
\newblock Mammoth2: Scaling instructions from the web.
\newblock \emph{arXiv preprint arXiv:2405.03548}.

\bibitem[{Zhang et~al.(2023)Zhang, Chen, Liu, and He}]{zhang2023composing}
Jinghan Zhang, Shiqi Chen, Junteng Liu, and Junxian He. 2023.
\newblock \href {https://openreview.net/forum?id=5r3e27I9Gy} {Composing parameter-efficient modules with arithmetic operation}.
\newblock In \emph{Thirty-seventh Conference on Neural Information Processing Systems}.

\bibitem[{Zhao et~al.(2024)Zhao, Li, Li, Zhang, and Sun}]{zhao-etal-2024-defending-large}
Wei Zhao, Zhe Li, Yige Li, Ye~Zhang, and Jun Sun. 2024.
\newblock \href {https://doi.org/10.18653/v1/2024.findings-emnlp.293} {Defending large language models against jailbreak attacks via layer-specific editing}.
\newblock In \emph{Findings of the Association for Computational Linguistics: EMNLP 2024}, pages 5094--5109, Miami, Florida, USA. Association for Computational Linguistics.

\bibitem[{Zhou et~al.(2024)Zhou, Song, Wang, and Chen}]{zhou-etal-2024-metagpt}
Yuyan Zhou, Liang Song, Bingning Wang, and Weipeng Chen. 2024.
\newblock \href {https://doi.org/10.18653/v1/2024.emnlp-main.102} {{M}eta{GPT}: Merging large language models using model exclusive task arithmetic}.
\newblock In \emph{Proceedings of the 2024 Conference on Empirical Methods in Natural Language Processing}, pages 1711--1724, Miami, Florida, USA. Association for Computational Linguistics.

\end{thebibliography}
\bibliographystyle{acl_natbib}

\appendix

\section{Appendix}
\label{sec:appendix}

\subsection{Models Used in Experiments}\label{appdix:models}
\subsubsection{Task Learning}
We show more details of models used for task learning when the model structure is Gemma-2-9b.\\
\textbf{Base Model:} gemma-2-9b\footnote{\href{https://huggingface.co/google/gemma-2-9b}{https://huggingface.co/google/gemma-2-9b}} \\
\textbf{Pre-Trained / Target Model:} gemma-2-9b-it\footnote{\href{https://huggingface.co/google/gemma-2-9b-it}{https://huggingface.co/google/gemma-2-9b-it}} \\
\textbf{Fine-Tuned Models:} \\
\textbf{UA:} gemma-2-9b-it-abliterated\footnote{\href{https://huggingface.co/IlyaGusev/gemma-2-9b-it-abliterated}{https://huggingface.co/IlyaGusev/gemma-2-9b-it-abliterated}} \\
\textbf{Math:} gemma-2-9b-it-mathinstruct\footnote{\href{https://huggingface.co/kyungeun/gemma-2-9b-it-mathinstruct}{https://huggingface.co/kyungeun/gemma-2-9b-it-mathinstruct}} \\
\textbf{Code:} gemma\_coder\_9b\footnote{\href{https://huggingface.co/TeamDelta/gemma_coder_9b}{https://huggingface.co/TeamDelta/gemma\_coder\_9b}}
\\
\\
We show more details of models used for task learning when the model structure is Llama-3-8B.\\
\textbf{Base Model:} Meta-Llama-3-8B\footnote{\href{https://huggingface.co/meta-llama/Meta-Llama-3-8B}{https://huggingface.co/meta-llama/Meta-Llama-3-8B}\label{llama-3-8b}} \\
\textbf{Pre-Trained / Target Model:} Meta-Llama-3-8B-Instruct\footnote{\href{https://huggingface.co/meta-llama/Meta-Llama-3-8B-Instruct}{https://huggingface.co/meta-llama/Meta-Llama-3-8B-Instruct}\label{llama-3-8b-instruct}} \\
\textbf{Fine-Tuned Models:} \\
\textbf{UA:} LLama-3-8b-Uncensored\footnote{\href{https://huggingface.co/DevsDoCode/LLama-3-8b-Uncensored}{https://huggingface.co/DevsDoCode/LLama-3-8b-Uncensored}\label{llama-UA}} \\
\textbf{Math:} MAmmoTH2-8B-Plus\footnote{\href{https://huggingface.co/TIGER-Lab/MAmmoTH2-8B-Plus}{https://huggingface.co/TIGER-Lab/MAmmoTH2-8B-Plus}} \\
\textbf{Code:} code-millenials-8b\footnote{\href{https://huggingface.co/budecosystem/code-millenials-8b}{https://huggingface.co/budecosystem/code-millenials-8b}}

\subsubsection{Task Forgetting}
We show more details of models used for task forgetting.
\\
\textbf{Base Model:} Meta-Llama-3-8B\textsuperscript{\ref{llama-3-8b}} \\
\textbf{Pre-Trained Model:} Meta-Llama-3-8B-Instruct\textsuperscript{\ref{llama-3-8b-instruct}} \\
\textbf{Fine-Tuned Models:} LLama-3-8b-Uncensored\textsuperscript{\ref{llama-UA}} \\
\textbf{Target Models:} \\
\textbf{Traditional Chinese:} Llama3-TAIDE-LX-8B-Chat-Alpha1\footnote{\href{https://huggingface.co/taide/Llama3-TAIDE-LX-8B-Chat-Alpha1}{https://huggingface.co/taide/Llama3-TAIDE-LX-8B-Chat-Alpha1}} \\
\textbf{German:} Llama3-DiscoLeo-Instruct-8B-v0.1\footnote{\href{https://huggingface.co/DiscoResearch/Llama3-DiscoLeo-Instruct-8B-v0.1}{https://huggingface.co/DiscoResearch/Llama3-DiscoLeo-Instruct-8B-v0.1}} \\
\textbf{Japanese:} Llama-3-ELYZA-JP-8B\footnote{\href{https://huggingface.co/elyza/Llama-3-ELYZA-JP-8B}{https://huggingface.co/elyza/Llama-3-ELYZA-JP-8B}} \\
\textbf{Russian:} saiga\_llama3\_8b\footnote{\href{https://huggingface.co/IlyaGusev/saiga_llama3_8b}{https://huggingface.co/IlyaGusev/saiga\_llama3\_8b}} \\
\textbf{Thai:} llama-3-typhoon-v1.5-8b-instruct\_8b\footnote{\href{https://huggingface.co/scb10x/llama-3-typhoon-v1.5-8b-instruct}{https://huggingface.co/scb10x/llama-3-typhoon-v1.5-8b-instruct}} \\

\subsection{Results with Different Hyperparameters on Gemma-2-9b}\label{appdix:results-gemma}
In this section, we show different hyperparameters of DARE, TIES, and DARE+TIES across different scaling coefficients on Gemma-2-9b. The results explain why the hyperparameters we used in the main text are the most effective for all baselines.
\paragraph{DARE.}
Table~\ref{table:8} follows the same settings as Table~\ref{table:5} while demonstrating the performance with varying drop rates. DARE achieves better results when the drop rate is 0.3.
\begin{table}[h]\tiny
\tabcolsep=3.5pt
\centering
\begin{tabular}{@{}cccccc@{}}
\hline
\begin{tabular}[c]{@{}c@{}}Merged\\ Tasks\end{tabular} & \begin{tabular}[c]{@{}c@{}}Drop \\ Rate\end{tabular} & \begin{tabular}[c]{@{}c@{}}Utility\\ WikiText-2($\downarrow$)\end{tabular} & \begin{tabular}[c]{@{}c@{}}UA\\ GPT-4($\uparrow$)\end{tabular} & \begin{tabular}[c]{@{}c@{}}Math\\ GSM8K($\uparrow$)\end{tabular} & \begin{tabular}[c]{@{}c@{}}Code\\ HumanEval($\uparrow$)\end{tabular} \\ \midrule
\hline
 & 0.3 & 10.0146 & 1.5909 & 0.8324 & - \\
UA + Math & 0.6 & 10.0979 & 1.6333 & 0.8241 & - \\
 & 0.9 & 10.5492 & 1.6242 & 0.8112 & - \\
\hline
 & 0.3 & 10.4055 & - & 0.8294 & 0.6341 \\
Math + Code & 0.6 & 10.4918 & - & 0.8393 & 0.6220 \\
 & 0.9 & 11.4782 & - & 0.7703 & 0.5671 \\
\hline
 & 0.3 & 10.8740 & 1.7061 & - & 0.4390 \\
UA + Code & 0.6 & 10.9795 & 1.6818 & - & 0.4634 \\
 & 0.9 & 11.3437 & 1.8303 & - & 0.5366 \\
\hline
 & 0.3 & 10.3804 & 1.6394 & 0.8294 & 0.6463 \\
UA + Math + Code & 0.6 & 10.4883 & 1.7091 & 0.8249 & 0.6280 \\
 & 0.9 & 11.6337 & 1.8636 & 0.7453 & 0.5305 \\
\hline
\end{tabular}
\caption{Results of task learning with DARE under Gemma-2-9b. All scaling coefficients here are set as $0.5$.}
\label{table:8}
\end{table}

On the other hand, we also consider different values of scaling coefficients. Following the settings of Table~\ref{table:6}, in Table~\ref{table:9}, we show the performance of DARE with different drop rates and the coefficient fixed at 1.0. Overall, compared with Table~\ref{table:6}, we obtain the best result for DARE when the drop rate is set to 0.3. This is why we choose these parameters in Table~\ref{table:5} and ~\ref{table:6}.

\begin{table}[h]\tiny
\tabcolsep=3.5pt
\centering
\begin{tabular}{@{}cccccc@{}}
\hline
\begin{tabular}[c]{@{}c@{}}Merged\\ Tasks\end{tabular} & \begin{tabular}[c]{@{}c@{}}Drop \\ Rate\end{tabular} & \begin{tabular}[c]{@{}c@{}}Utility\\ WikiText-2($\downarrow$)\end{tabular} & \begin{tabular}[c]{@{}c@{}}UA\\ GPT-4($\uparrow$)\end{tabular} & \begin{tabular}[c]{@{}c@{}}Math\\ GSM8K($\uparrow$)\end{tabular} & \begin{tabular}[c]{@{}c@{}}Code\\ HumanEval($\uparrow$)\end{tabular} \\ \midrule
\hline
 & 0.3 & 10.8753 & 3.9424 & 0.7437 & - \\
UA + Math & 0.6 & 11.2912 & 3.8515 & 0.7036 & - \\
 & 0.9 & 15.3662 & 3.7636 & 0.4594 & - \\
\hline
 & 0.3 & 12.4347 & - & 0.7111 & 0.5305 \\
Math + Code & 0.6 & 13.2541 & - & 0.6626 & 0.4329 \\
 & 0.9 & 49.7530 & - & 0.0205 & 0.0183 \\
\hline
 & 0.3 & 12.1404 & 3.8394 & - & 0.3537 \\
UA + Code & 0.6 & 12.3582 & 3.7697 & - & 0.3354 \\
 & 0.9 & 16.0632 & 3.3121 & - & 0.3171 \\
\hline
 & 0.3 & 12.5596 & 3.6939 & 0.7005 & 0.5061 \\
UA + Math + Code & 0.6 & 13.5538 & 3.6061 & 0.6262 & 0.3902 \\
 & 0.9 & 61.5858 & \XSolid & 0.0091 & 0.0183 \\
\hline
\end{tabular}
\caption{Results of task learning with DARE under Gemma-2-9b. All scaling coefficients here are set as $1.0$. The cross sign indicates that the model can only generate gibberish.}\label{table:9}
\end{table}
\paragraph{TIES.} In Table~\ref{table:10}, we follow the same settings with Table~\ref{table:5}, but show more results for different $k$ of TIES. TIES obtain better utilities across different combinations of task merging when $k=0.7$.
\begin{table}[h]\tiny
\tabcolsep=3.5pt
\centering
\begin{tabular}{@{}cccccc@{}}
\hline
\begin{tabular}[c]{@{}c@{}}Merged\\ Tasks\end{tabular} & Top $k$ & \begin{tabular}[c]{@{}c@{}}Utility\\ WikiText-2($\downarrow$)\end{tabular} & \begin{tabular}[c]{@{}c@{}}UA\\ GPT-4($\uparrow$)\end{tabular} & \begin{tabular}[c]{@{}c@{}}Math\\ GSM8K($\uparrow$)\end{tabular} & \begin{tabular}[c]{@{}c@{}}Code\\ HumanEval($\uparrow$)\end{tabular} \\ \midrule
\hline
 & 0.5 & 10.0067 & 1.5394 & 0.8347 & - \\
UA + Math & 0.7 & 10.0167 & 1.5636 & 0.8385 & - \\
 & 0.9 & 10.0267 & 1.5788 & 0.8340 & - \\
\hline
 & 0.5 & 10.3486 & - & 0.8317 & 0.6707 \\
Math + Code & 0.7 & 10.3763 & - & 0.8279 & 0.6524 \\
 & 0.9 & 11.3946 & - & 0.8309 & 0.6524 \\
\hline
 & 0.5 & 10.8511 & 1.5455 & - & 0.4451 \\
UA + Code & 0.7 & 10.8583 & 1.5121 & - & 0.4329 \\
 & 0.9 & 10.8495 & 1.5455 & - & 0.4390 \\
\hline
 & 0.5 & 10.3727 & 1.6515 & 0.8264 & 0.6585 \\
UA + Math + Code & 0.7 & 10.3994 & 1.7939 & 0.8317 & 0.6585 \\
 & 0.9 & 10.4196 & 1.8636 & 0.8309 & 0.6463 \\
\hline
\end{tabular}
\caption{Results of task learning with TIES under Gemma-2-9b. All scaling coefficients here are set as $0.5$.}
\label{table:10}
\end{table}

Apart from the hyperparameter of TIES, we also take the scaling coefficient into account. Therefore, Table~\ref{table:11} uses the same settings as Table~\ref{table:6}, with the only difference being the top $k$. In comparison with Table~\ref{table:6}, the results are better when $k$ is 0.7. Therefore, the proper hyperparameters of TIES are setting $k$ as 0.7.

\begin{table}[h]\tiny
\tabcolsep=3.5pt
\centering
\begin{tabular}{@{}cccccc@{}}
\hline
\begin{tabular}[c]{@{}c@{}}Merged\\ Tasks\end{tabular} & Top $k$ & \begin{tabular}[c]{@{}c@{}}Utility\\ WikiText-2($\downarrow$)\end{tabular} & \begin{tabular}[c]{@{}c@{}}UA\\ GPT-4($\uparrow$)\end{tabular} & \begin{tabular}[c]{@{}c@{}}Math\\ GSM8K($\uparrow$)\end{tabular} & \begin{tabular}[c]{@{}c@{}}Code\\ HumanEval($\uparrow$)\end{tabular} \\ \midrule
\hline
 & 0.5 & 10.6077 & 3.3455 & 0.7635 & - \\
UA + Math & 0.7 & 10.7638 & 3.3606 & 0.7475 & - \\
 & 0.9 & 10.8087 & 3.5394 & 0.7362 & - \\
\hline
 & 0.5 & 11.9588 & - & 0.7460 & 0.5732 \\
Math + Code & 0.7 & 12.3455 & - & 0.7165 & 0.5366 \\
 & 0.9 & 12.5847 & - & 0.6892 & 0.5427 \\
\hline
 & 0.5 & 11.8724 & 3.4182 & - & 0.3049 \\
UA + Code & 0.7 & 11.9742 & 3.3545 & - & 0.2866 \\
 & 0.9 & 11.9892 & 3.4455 & - & 0.2927 \\
\hline
 & 0.5 & 12.1112 & 3.3848 & 0.7263 & 0.5732 \\
UA + Math + Code & 0.7 & 12.5602 & 3.5061 & 0.7104 & 0.5366 \\
 & 0.9 & 12.8162 & 3.5727 & 0.6907 & 0.5244 \\
\hline
\end{tabular}
\caption{Results of task learning with TIES under Gemma-2-9b. All scaling coefficients here are set as $1.0$.}
\label{table:11}
\end{table}
\paragraph{DARE + TIES.} Here, we show more different hyperparameter combinations of DARE+TIES with the scaling coefficient set to 0.5 in Table~\ref{table:12}. Across different tasks, the results with ($p$, $k$) = (0.1, 0.9) outperform other settings.  These are the parameters we use in the main text as well.
\begin{table}[t]\tiny
\tabcolsep=1.5pt
\centering
\begin{tabular}{@{}cccccc@{}}
\hline
\begin{tabular}[c]{@{}c@{}}Merged\\ Tasks\end{tabular} & \begin{tabular}[c]{@{}c@{}}Drop Rate $p$\\/ Top $k$\end{tabular} & \begin{tabular}[c]{@{}c@{}}Utility\\ WikiText-2($\downarrow$)\end{tabular} & \begin{tabular}[c]{@{}c@{}}UA\\ GPT-4($\uparrow$)\end{tabular} & \begin{tabular}[c]{@{}c@{}}Math\\ GSM8K($\uparrow$)\end{tabular} & \begin{tabular}[c]{@{}c@{}}Code\\ HumanEval($\uparrow$)\end{tabular} \\ \midrule
\hline
 & 0.7 / 0.3 & 10.1749 & 1.6000 & 0.8127 & - \\
UA + Math & 0.4 / 0.6 & 10.0813 & 1.5545 & 0.8332 & - \\
 & 0.1 / 0.9 & 10.0461 & 1.5818 & 0.8302 & - \\
\hline
 & 0.7 / 0.3 & 10.7264 & - & 0.8173 & 0.6341 \\
Math + Code & 0.4 / 0.6 & 10.4415 & - & 0.8294 & 0.6524 \\
 & 0.1 / 0.9 & 10.4208 & - & 0.8287 & 0.6341 \\
\hline
 & 0.7 / 0.3 & 10.9549 & 1.6091 & - & 0.4329 \\
UA + Code & 0.4 / 0.6 & 10.8592 & 1.6030 & - & 0.4512 \\
 & 0.1 / 0.9 & 10.8553 & 1.6121 & - & 0.4512 \\
\hline
 & 0.7 / 0.3 & 10.7478 & 1.7333 & 0.8089 & 0.6280 \\
UA + Math + Code & 0.4 / 0.6 & 10.4725 & 1.7727 & 0.8279 & 0.6646 \\
 & 0.1 / 0.9 & 10.4147 & 1.8091 & 0.8309 & 0.6524 \\
\hline
\end{tabular}
\caption{Results of task learning with DARE + TIES under Gemma-2-9b. All scaling coefficients here are set as $0.5$.}
\label{table:12}
\end{table}

\subsection{Results with Different Hyperparameters on Llama-3-8B}\label{appdix:results-llama}
In this section, we present various hyperparameters for DARE, TIES, and DARE+TIES on Llama-3-8B. The results demonstrate why the hyperparameters chosen in the main text are the most optimal across all baselines.
\paragraph{DARE.} In the main text, we show the results of DARE when the drop rate is 0.3 and the scaling coefficient is 0.5 on the Llama-3-8B model. Table~\ref{table:13} presents additional results of DARE using different drop rate settings. However, Table~\ref{table:13} demonstrates that DARE can get the best result when the drop rate is set as 0.3.
\begin{table}[h]\tiny
\tabcolsep=3.5pt
\centering
\begin{tabular}{@{}cccccc@{}}
\hline
\begin{tabular}[c]{@{}c@{}}Merged\\ Tasks\end{tabular} & \begin{tabular}[c]{@{}c@{}}Drop \\ Rate\end{tabular} & \begin{tabular}[c]{@{}c@{}}Utility\\ WikiText-2($\downarrow$)\end{tabular} & \begin{tabular}[c]{@{}c@{}}UA\\ GPT-4($\uparrow$)\end{tabular} & \begin{tabular}[c]{@{}c@{}}Math\\ GSM8K($\uparrow$)\end{tabular} & \begin{tabular}[c]{@{}c@{}}Code\\ HumanEval($\uparrow$)\end{tabular} \\ \midrule
\hline
 & 0.3 & 9.0559 & 3.5606 & 0.8074 & - \\
UA + Math & 0.6 & 9.2150 & 3.6576 & 0.7900 & - \\
 & 0.9 & 10.3136 & 3.3394 & 0.7195 & - \\
\hline
 & 0.3 & 10.1674 & - & 0.6558 & 0.3415 \\
Math + Code & 0.6 & 10.5052 & - & 0.6626 & 0.2195 \\
 & 0.9 & 14.0010 & - & 0.4814 & 0.1707 \\
\hline
 & 0.3 & 10.4840 & 3.7273 & - & 0.2256 \\
UA + Code & 0.6 & 10.6566 & 3.6303 & - & 0.1463 \\
 & 0.9 & 11.6871 & 3.7939 & - & 0.1646 \\
\hline
 & 0.3 & 10.0415 & 3.8000 & 0.6732 & 0.3171 \\
UA + Math + Code & 0.6 & 10.2933 & 3.5061 & 0.6467 & 0.2866 \\
 & 0.9 & 13.3988 & 3.9152 & 0.4723 & 0.1159 \\
\hline
\end{tabular}
\caption{Results of task learning with DARE under Llama-3-8B. All scaling coefficients here are set as $0.5$.}
\label{table:13}
\end{table}

\paragraph{TIES.} Fixing the scaling coefficient at 0.5, we conduct more experiments of TIES on Llama-3-8B for different values of $k$, and results are shown in Table~\ref{table:14}. Most of results with $k=7$ surpass the other values of $k$.
\begin{table}[h]\tiny
\tabcolsep=3.5pt
\centering
\begin{tabular}{@{}cccccc@{}}
\hline
\begin{tabular}[c]{@{}c@{}}Merged\\ Tasks\end{tabular} & Top $k$ & \begin{tabular}[c]{@{}c@{}}Utility\\ WikiText-2($\downarrow$)\end{tabular} & \begin{tabular}[c]{@{}c@{}}UA\\ GPT-4($\uparrow$)\end{tabular} & \begin{tabular}[c]{@{}c@{}}Math\\ GSM8K($\uparrow$)\end{tabular} & \begin{tabular}[c]{@{}c@{}}Code\\ HumanEval($\uparrow$)\end{tabular} \\ \midrule
\hline
 & 0.5 & 9.1528 & 3.3788 & 0.8036 & - \\
UA + Math & 0.7 & 9.0490 & 3.4909 & 0.8089 & - \\
 & 0.9 & 9.0009 & 3.6636 & 0.7983 & - \\
\hline
 & 0.5 & 10.0103 & - & 0.6914 & 0.3293 \\
Math + Code & 0.7 & 10.1618 & - & 0.6831 & 0.3354 \\
 & 0.9 & 10.2093 & - & 0.6732 & 0.3232 \\
\hline
 & 0.5 & 10.2491 & 3.5667 & - & 0.2256 \\
UA + Code & 0.7 & 10.4020 & 3.6818 & - & 0.1646 \\
 & 0.9 & 10.5076 & 3.6727 & - & 0.1707 \\
\hline
 & 0.5 & 9.9066 & 3.5030 & 0.6793 & 0.3171 \\
UA + Math + Code & 0.7 & 10.0566 & 3.5697 & 0.6694 & 0.2622 \\
 & 0.9 & 10.1106 & 3.5818 & 0.6535 & 0.3171 \\
\hline
\end{tabular}
\caption{Results of task learning with TIES under Llama-3-8B. All scaling coefficients here are set as $0.5$.}
\label{table:14}
\end{table}

\paragraph{DARE+TIES.}
We run more experiments of DARE+TIES on Llama-3-8B to show the impacts of different combinations of the drop rate and top $k$. Table~\ref{table:15} shows the results, with (p, k) = (0.1, 0.9) achieving the best performance in most cases. This indicates that the parameters we use in the main text are the most favorable for this method.
\begin{table}[h]\tiny
\tabcolsep=1.5pt
\centering
\begin{tabular}{@{}cccccc@{}}
\hline
\begin{tabular}[c]{@{}c@{}}Merged\\ Tasks\end{tabular} & \begin{tabular}[c]{@{}c@{}}Drop Rate $p$\\/ Top $k$\end{tabular} & \begin{tabular}[c]{@{}c@{}}Utility\\ WikiText-2($\downarrow$)\end{tabular} & \begin{tabular}[c]{@{}c@{}}UA\\ GPT-4($\uparrow$)\end{tabular} & \begin{tabular}[c]{@{}c@{}}Math\\ GSM8K($\uparrow$)\end{tabular} & \begin{tabular}[c]{@{}c@{}}Code\\ HumanEval($\uparrow$)\end{tabular} \\ \midrule
\hline
 & 0.7 / 0.3 & 9.3173 & 3.7091 & 0.7983 & - \\
UA + Math & 0.4 / 0.6 & 9.0607 & 3.5471 & 0.7945 & - \\
 & 0.1 / 0.9 & 9.0055 & 3.5515 & 0.7923 & - \\
\hline
 & 0.7 / 0.3 & 10.8194 & - & 0.6391 & 0.2683 \\
Math + Code & 0.4 / 0.6 & 10.3354 & - & 0.6520 & 0.3293 \\
 & 0.1 / 0.9 & 10.2170 & - & 0.6778 & 0.2927 \\
\hline
 & 0.7 / 0.3 & 10.8128 & 3.6030 & - & 0.0976 \\
UA + Code & 0.4 / 0.6 & 10.5554 & 3.7242 & - & 0.2256 \\
 & 0.1 / 0.9 & 10.5172 & 3.6606 & - & 0.1951 \\
\hline
 & 0.7 / 0.3 & 10.7319 & 3.8182 & 0.6224 & 0.3232 \\
UA + Math + Code & 0.4 / 0.6 & 10.2063 & 3.6303 & 0.6535 & 0.3293 \\
 & 0.1 / 0.9 & 10.1180 & 3.6727 & 0.6634 & 0.3537 \\
\hline
\end{tabular}
\caption{Results of task learning with DARE + TIES under Llama-3-8B. All scaling coefficients here are set as $0.5$.}
\label{table:15}
\end{table}

\end{document}